\documentclass{article}

\usepackage{arxiv}
\usepackage{ulem}
\usepackage{subfig}
\usepackage{soul,color}
\usepackage{graphicx}
\usepackage{changepage}
\usepackage[utf8]{inputenc} 
\usepackage[T1]{fontenc}    
\usepackage{hyperref}       
\usepackage{url}            
\usepackage{booktabs}       
\usepackage{amsfonts}       
\usepackage{nicefrac}       
\usepackage{microtype}      
\usepackage{lipsum}
\usepackage[table]{xcolor}
\usepackage{caption} 
\usepackage{float}
\usepackage[font={small,bf}]{caption}
\usepackage{xcolor}

\usepackage{natbib}

\definecolor{nkGreen}{RGB}{0,102,102}

\usepackage{array}
\usepackage{graphicx}
\usepackage{multirow}

\usepackage{ctable}
\title{Applications of deep learning in congestion detection, prediction and alleviation: A survey}

\author{Nishant Kumar \\
  ETH Zurich\\
  Future Resilient Systems\\
  Singapore-ETH Centre\\
  1 CREATE Way\\
  \#06-01 CREATE Tower\\
  Singapore-138602\\
  \texttt{nishant.kumar@sec.ethz.ch} 
   \And
 Martin Raubal\\
 ETH Zurich\\
  Department of Civil,\\ Environmental and Geomatic Engineering \\
  Stefano-Franscini-Platz 5\\
   8093 Zurich\\
   Switzerland\\
  \texttt{mraubal@ethz.ch} \\
}

\usepackage[percent]{overpic}
\def\arrvline{\hfil\kern\arraycolsep\vline\kern-\arraycolsep\hfilneg}

\begin{document}
\maketitle

\begin{abstract}
Detecting, predicting, and alleviating traffic congestion are targeted at improving the level of service of the transportation network. With increasing access to larger datasets of higher resolution, the relevance of deep learning for such tasks is increasing. Several comprehensive survey papers in recent years have summarised the deep learning applications in the transportation domain. However, the system dynamics of the transportation network vary greatly between the non-congested state and the congested state -- thereby necessitating the need for a clear understanding of the challenges specific to congestion prediction. In this survey, we present the current state of deep learning applications in the tasks related to detection, prediction, and alleviation of congestion. Recurring and non-recurring congestion are discussed separately. Our survey leads us to uncover inherent challenges and gaps in the current state of research. Finally, we present some suggestions for future research directions as answers to the identified challenges. 
\end{abstract}

\section{Introduction}
Traffic congestion decreases the level of serviceability  (LOS) of road networks. A decrease in LOS results in direct and indirect costs to society. Extensive studies have been carried out to estimate the impacts of congestion on the  economy and society as a whole \citep{weisbrod2001economic,litman2016smart}. The first-hand impact of traffic congestion is the lost working hours.~\cite{schrank20122012} estimated that in a single year, the USA alone lost a total of 8.8 billion working hours due to congestion. The detrimental impacts of congestion skyrocket when the value of time, as a commodity, increases drastically during emergencies. Being stuck in traffic impacts the behaviour of individuals.~\cite{hennessy1999traffic} report that high congestion levels can result in aggressive behaviour by drivers. This aggression can manifest itself into aggressive driving, thereby increasing the chances of accidents~\citep{li2020influence}. High levels of congestion also result in higher greenhouse gas emissions~\citep{barth2009traffic}. \\


In terms of tractability, congestion prediction is a more difficult problem than traffic prediction during uncongested conditions~\citep{yu2017deep}. An early warning system enables traffic controllers to put alleviation measures in place. The infrastructure required for traffic data collection has improved over the decades. This improvement, in conjunction with the increased availability of computational resources, has enabled transportation researchers to leverage the predictive capabilities of deep neural networks for this domain. In this survey, we discuss the applications of deep learning in the detection, prediction, and alleviation of congestion. We investigate various aspects of the two types of congestion -- recurring and non-recurring. Towards the end of this survey, we identify some gaps in the current state of the research in this field and present future research directions. 
\\

\section{Preliminaries}
The target audience of this survey paper are  researchers from two backgrounds- transportation and deep learning. In the following two subsections, we cover the preliminaries and introduce the terminology which is used throughout the survey.  



\subsection{Relevant concepts and terms in deep learning}\label{section:machineTrans}
An artificial neuron is a function as shown in~\autoref{equation1}
\begin{equation}
y_{i}=\varphi\left (\sum_{j=0}^{m} w_{ij} x_{ij}\right)
\label{equation1}
\end{equation}
where $x_{ij}$ is the $j^{th}$ feature (dimension) of the $i^{th}$ $m$-dimensional data point in the dataset; $w_{j}$ (called weights) is the coefficient which is tuned during the training process of the neural network;  $\varphi$ is a nonlinear activation function; $y_i$ is the output of the function on input $x_i$. Commonly used activation functions are: sigmoid ($\varphi (x)=\frac{1}{1+e^{-x}}$), tanh ($\varphi (x)=\frac{e^{2 x}-1}{e^{2 x}+1}$) and relu ($\varphi (x)=max (x,0)$). \\

\begin{figure}[h]
    \centering
    \includegraphics[width=0.7\textwidth]{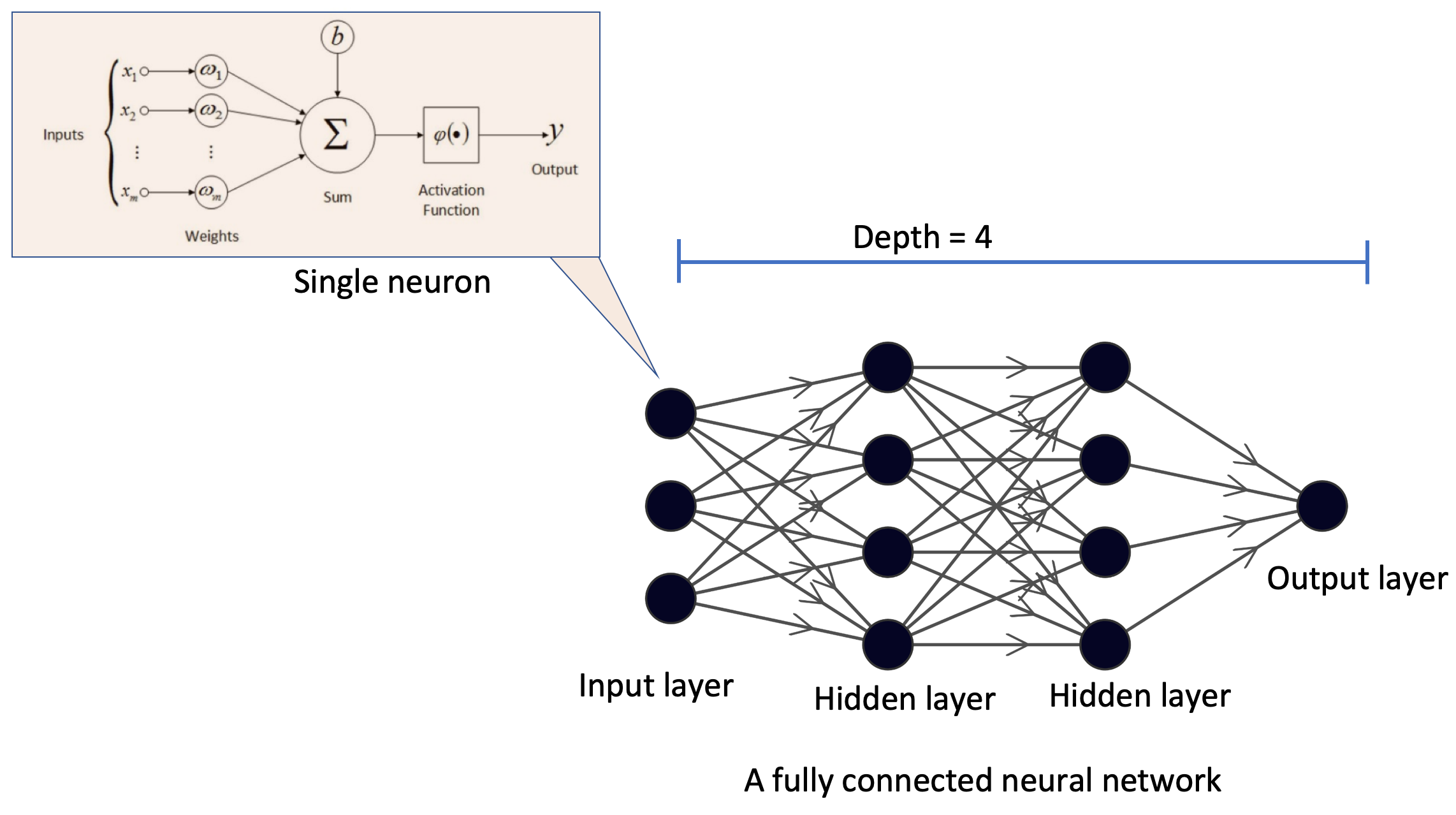}
    \captionsetup{margin=2cm}
  \caption{\texttt{A fully connected layered neural network with two hidden layers. The term `layered' is often omitted in the literature. In this survey, we refer to these as feed forward neural networks. An enlarged diagram of a single artificial neuron is presented separately to show its three components -- inputs, weighted sum, and nonlinear activation. The input data point has $n$ features (also called dimensions). Image sources: Single neuron image obtained from Chrislb - Own work, \href{https://commons.wikimedia.org/w/index.php?curid=224550}{\textcolor{blue}{wikimedia public domain CC BY-SA 3.0)}}. The layered network created using Nn-svg~\citep{lenail2019nn} }} 
    \label{nnwalabasicfigure}
\end{figure}

\subsubsection{Fully connected layered neural networks} The network shown in~\autoref{nnwalabasicfigure} is a fully connected layered neural network. In such neural networks, the outputs from all neurons from a previous layer are fed as inputs to all neurons in the next layer. The literature on neural networks often omits the term `layered' and uses the terms fully connected neural network (FCNN) in its place. The usage can be a source of confusion because a fully connected network might imply the presence of connections between all neurons in the neural network, not just between layers placed next to each other. In order to ensure proper terminology, in this survey we stick to the term feed forward neural network (FFNN) to imply fully connected layered neural networks~\citep[Chapter~6]{goodfellow2016deep}. Additionally, we observed that several papers referred to in this survey used the term artificial neural network (ANN) to denote FFNN. In this survey, however, we stick to FFNN and avoid using ANN in its place. \\

When placed parallel to each other, several such neurons form a \textit{layer} of the neural network. When several layers are stacked one after the other, a feed forward neural network (FFNN) is formed. In this context, stacking refers to passing the output of one function or unit to another. The formation of a neural network using neurons is shown in~\autoref{nnwalabasicfigure}. As we increase the stacking, the \textit{depth} of the neural network increases. The literature on neural networks does not specify a pre-defined threshold for the {depth}, in order to demarcate \textit{deep} and \textit{shallow} networks. Any neural network having more than one hidden layer can be referred to as a {deep} neural network~\citep{schmidhuber2015deep}. A {deep} neural network can learn more abstract representations of the data compared to a shallow network. The link between depth and abstractions is easily observed when working with image data as shown in~\autoref{cnnwalafigure}. We shall frequently use the terms {depth} and {layers} in this survey. \\

For supervised learning tasks using deep learning, such as the prediction of congestion, the goal is to train a deep learning model so that it learns a mapping from input data to the output data. Let us consider a traffic prediction task where the goal is to predict the traffic flow at $n$ locations 1 minute into the future. The input data $x$ is a vector of length $n$ and varies at every minute $t$. Training the deep learning model implies assuming the existence of an underlying function $f$ such that $x(t+1) = f(x(t))$ and then, attempt to approximate $f$ by adapting the weights of the model. In order to approximate the function $f(x)$, a loss function $L$ is minimised. The most commonly used loss function is mean squared error. The minimisation is carried out using the backpropagation algorithm~\citep{BPNN}. Backpropagation refers to the propagation of errors from the loss function to the previous layers using the chain rule of differentiation. In order to  guarantee that the function approximated by the deep learning model is not arbitrary, the loss is computed on new unseen data (test data) after the training process. If the values of loss function on training data and test data are similar, the model is said to generalise well. Several techniques for better generalisation of deep learning models have been explored. The most commonly used technique is dropout~\citep{srivastava2014dropout}. In this survey, we pay special attention to the generalisation techniques which have been developed by  leveraging upon the domain knowledge from transportation. 

\subsubsection{CNNs and RNNs} Traffic data vary over space and time. Two families of neural network architectures are particularly suited to capture such inter-dependencies: CNN and RNN.\\

\begin{figure}[ht]
    \centering
    \includegraphics[width=0.9\textwidth]{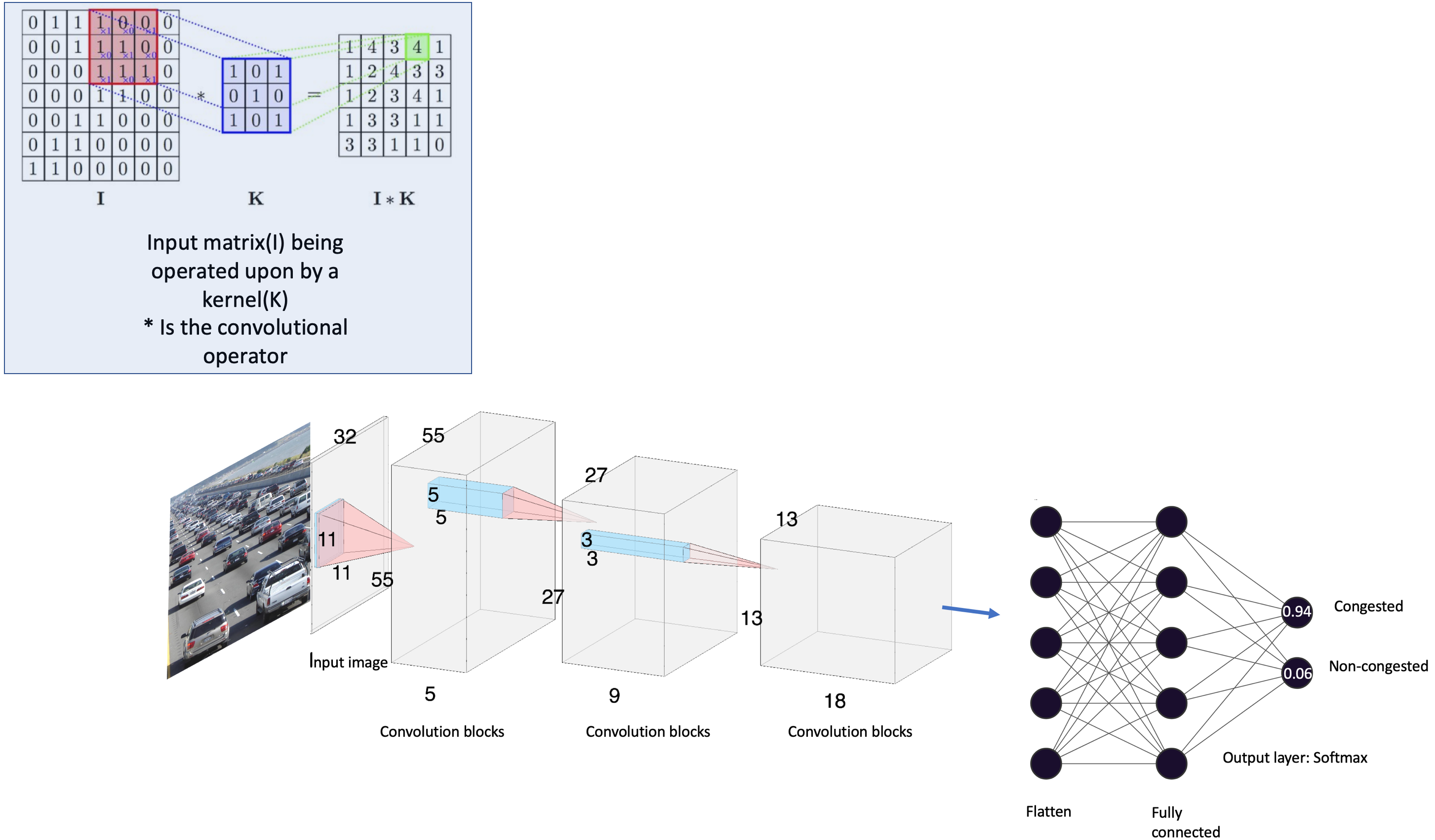}
    \captionsetup{margin=2cm}
    \caption{\texttt{ (Top: A single convolution operation; Bottom: A deep convolutional neural network (CNN). As data propagate through various convolutional layers, the features learnt become more high level. Image sources: single convolution image adapted from~\citep{mohamed2017detection} with permission, traffic jam image adapted from \href{https://upload.wikimedia.org/wikipedia/commons/3/3e/I-80_Eastshore_Fwy.jpg
}{\textcolor{blue}{wikimedia public  domain CC BY-SA 3.0)} }}}
    \label{cnnwalafigure}
\end{figure}

\textbf{Convolutional Neural Networks:} CNN stands for Convolutional Neural Network. Historically, CNNs have been popularly used in image-classification problems due to their ability to capture the correlation between nearby pixels of an image. A deep-CNN is able to capture the correlations between pixels placed far apart in the image. In a typical CNN architecture, the first few layers are convolutional blocks, interspersed with pooling layers. Fully connected layers are present just before the output layer. Pooling is a downsampling technique used to report summary statistics from a neighbourhood~\citep[chap.~9]{goodfellow2016deep}. The most commonly used pooling method with CNNs is max-pooling, wherein the maximum value of the activation is selected from a region~\citep{maxpoolcitation}. Pooling helps reduce the complexity of the deep learning model and also learn representations that are invariant to small local translations of the input data. The effectiveness of deep CNNs in capturing the spatial dependencies in the image is illustrated in~\autoref{cnnwalafigure}. When applied to traffic data such as a 2-D image of grid-wise congestion level, the CNN models can capture the spatial dependencies.\\

\begin{figure}
\centering
{\includegraphics[width=13cm]{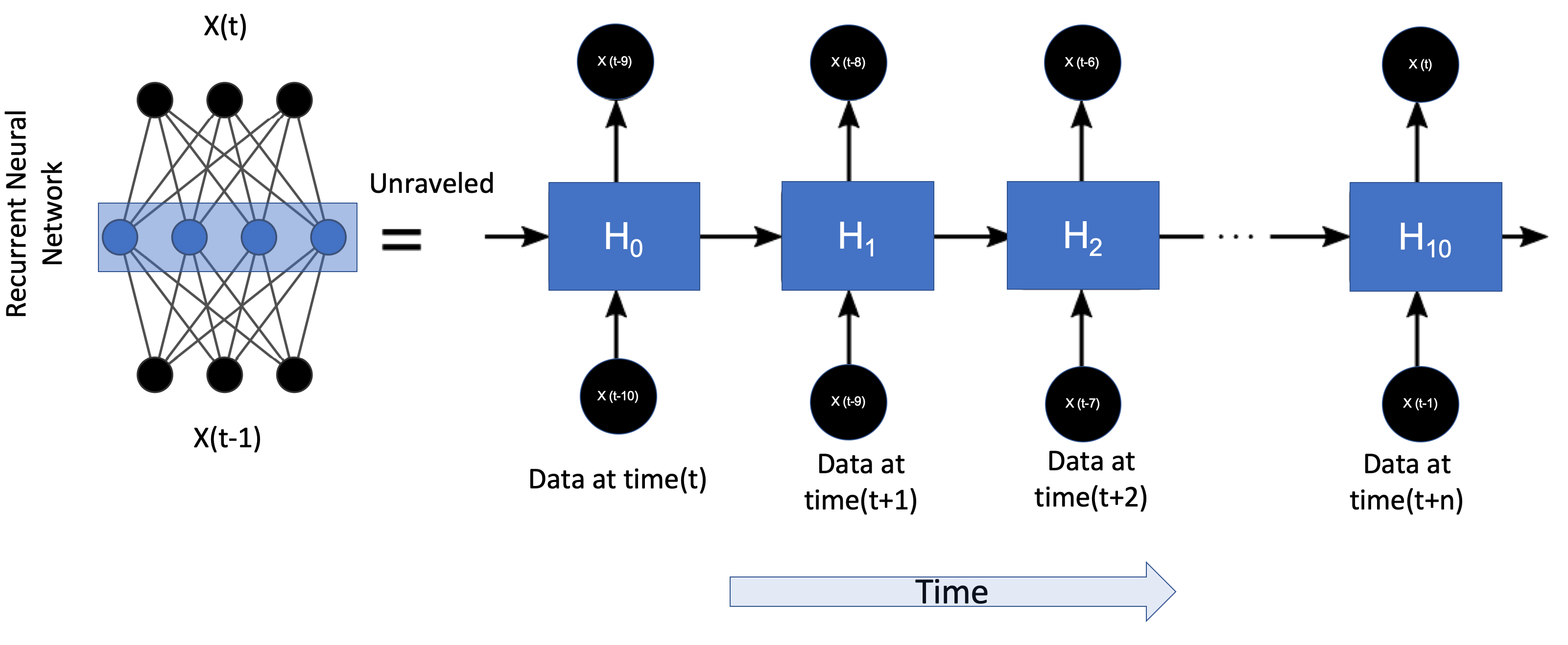}}

\captionsetup{margin=2cm}
\caption{\texttt{\texttt{Unraveling of a recurrent neural network in time}}} 
\label{rnnlstmwala} 
\end{figure}

\textbf{Recurrent Neural Networks:} Based on the previous example, if we extend our assumed underlying function $f$ to be dependent on the previous 10 time steps instead of just one, ($x(t+1) = f(x(t), x(t-1), x(t-2), x(t-3),..x(t-10))$, training an FFNN can be achieved by concatenating sequences of 10 input vectors to create input vectors ($x'(t)$) of length ($n*10$). The other option is to do backpropagation through time. This is accomplished using Recurrent Neural Networks (RNNs) as shown in~\autoref{rnnlstmwala}. A RNN has a feedback loop in the connection, as opposed to the feed forward neural networks. The feedback loop can be unravelled to uncover the back propagation in time. When time series data are passed in sequence as an input to a RNN, the RNN maintains an internal state from one time-step to the next. At time $t+1$, the hidden state is influenced by the the input at time $t$ and the previous hidden state. This helps the RNNs to unravel the temporal dependencies in the data. Long-short-term-memory (LSTM) introduced in~\cite{lstm-main-paper}, is an improvement over the traditional RNN. LSTM networks can detect dependencies between  data points which are far apart in time~\citep{greff2016lstm}. At time $t$, an LSTM cell is characterised by the state of four logic gates- input ($i_t$), output ($o_t$), cell state ($c_t$) and forget ($f_t$) gates. Using our example of the vector $x(t)$, the hidden state ($h_t$) of an LSTM unit can be formalised using equations  2-6~\citep{lstm2,WANG2019144}.\\


\small
\begin{equation} \label{lstm_eq1} i_{t} =\sigma\left(W_{\mathrm{xi}}x_{t}+W_{\mathrm{hi}} h_{t-1}\right) \end{equation}
\begin{equation}\label{lstm_eq3} f_{t} =\sigma\left(W_{\mathrm{xf}}x_{t}+W_{\mathrm{hf}}h_{t-1}\right) \end{equation}
\begin{equation}\label{lstm_eq4} o_{t} =\tanh \left(W_{\mathrm{xo}}x_{t}+W_{\mathrm{ho}} h_{t-1}\right) \end{equation} 
\begin{equation}\label{lstm_eq5} c_{t} =c_{t-1} \odot f_{t}+i_{t} \odot \tanh\left(W_{\mathrm{xc}}x_{t}+W_{\mathrm{hc}} h_{t-1}\right) \end{equation}
\begin{equation}\label{lstm_eq6} h_{t} =\tanh \left(c_{t}\right) \odot o_{t}  \end{equation}

\normalsize

where the $W_{ab}$ refers to the weight matrix between gates $a$ and $b$, $\odot$ refers to element-wise vector product, $h_t$ refers to the hidden state at time $t$, $i_t$ refers to the input at time $t$, $o_t$ refers to the output at time $t$ and $\sigma$ refers to the sigmoid activation function.

Another recent improvement over LSTM are the Gated Recurrent Units, proposed in~\cite{chung2014empirical}. Compared to LSTM, GRU has a less complex structure and can be trained faster than LSTM. At time $t$, a GRU cell is characterised by the state of two logic gates- update gate ($z_t$) and reset gate ($r_t$). For a detailed empirical comparison description of the differences between RNNs and LSTMs, the interested reader is referred to~\cite{jozefowicz2015empirical} for an empirical evaluation of GRUs and LSTMs. The hidden state of a GRU can be formalised using equations 7-9~\citep{WANG2019144}:
\small
 \begin{equation} r_{t}=\sigma\left(W_{\mathrm{xr}} x_{t}+W_{\mathrm{hr}} h_{t-1}\right) \end{equation}
 \begin{equation}  z_{t}=\sigma\left(W_{\mathrm{xz}} x_{t}+W_{\mathrm{hz}} h_{t-1}\right) \end{equation} 
 \begin{equation} h_{t}=z_{t} \odot h_{t-1}+\left(1-z_{t}\right) \odot \left( \tanh \left(W_{\mathrm{xh}} x_{t}+W_{\mathrm{hh}}\left(r_{t} \odot h_{t-1}\right)\right)\right) \end{equation}
 \normalsize
where the $W_{ab}$ refers to the weight matrix between gates $a$ and $b$, $\odot$ refers to element-wise vector product, $h_t$ refers to the hidden state at time $t$ and $\sigma$ refers to the sigmoid activation function.


\subsubsection{Reinforcement Learning} 
\begin{figure}[h]
    \centering
    \includegraphics[width=0.3\textwidth]{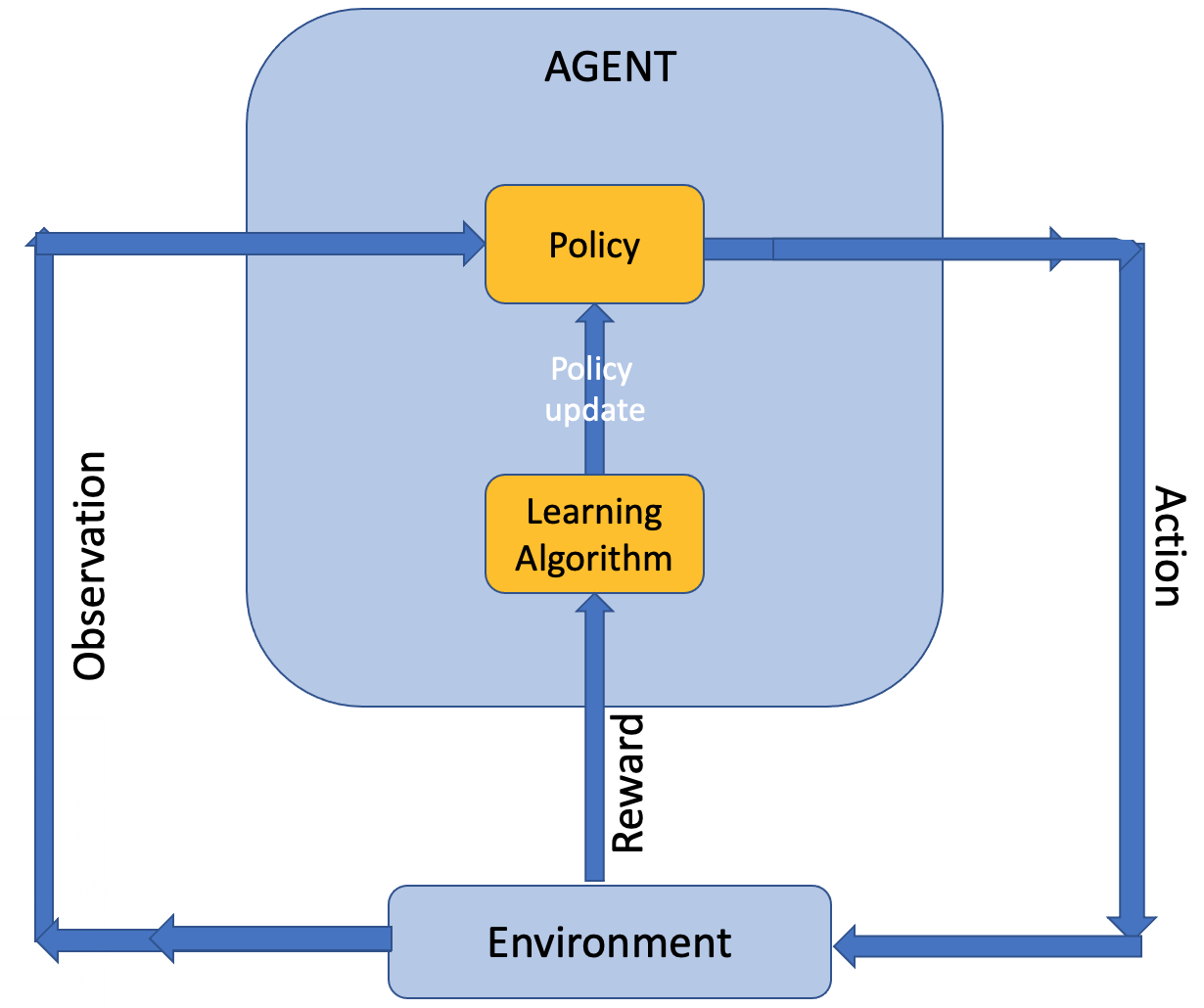}
  
    \captionsetup{margin=1cm}
    \caption{\texttt{The general framework of reinforcement learning}} 
    \label{reinforemcentll}
\end{figure}

Apart from CNNs and RNNs, a commonly used deep learning framework for traffic prediction tasks is deep reinforcement learning. Reinforcement learning (RL) is a learning paradigm that, when combined with deep learning, serves as a powerful tool for specific traffic-related prediction tasks where \textit{control} is involved. Deep-reinforcement learning models have been demonstrated to perform very well for specific tasks -- the most notable example is the model that was able to learn the game of \textit{Go} starting from scratch (basic game rules) to a level that surpassed the rating of world champions \citep{silver2017mastering}. The high computational load of such models, however, restricts their wide use.~\cite[Page-1, abstract]{szepesvari2010algorithms} define reinforcement learning as ``a learning paradigm concerned with learning \textit{to control} a system so as to maximize a numerical performance measure that expresses a long-term objective''. In this context, \textit{how to control} is also referred to as the \textit{policy} being learnt. When a deep learning model is trained to learn the best {policy}, it is referred to as a deep-reinforcement learning model. A representation of the reinforcement learning framework is presented in~\autoref{reinforemcentll}. While reviewing the literature, we found that Q-learning appears to be the popular reinforcement learning framework for traffic prediction tasks. Q-learning, introduced in~\cite{watkins1989learning}, is a model-free reinforcement learning approach where the \textit{environment} as shown in~\autoref{reinforemcentll}, does not need to be modelled explicitly. A detailed discussion on DQN is presented in~\cite{mnih2015human}.


\subsubsection{Commonly used metrics} 
The commonly used metrics to quantify performance for regression tasks are Mean Absolute Error (MAE), Mean Absolute Percentage Error (MAPE) and Root Mean Squared Error (RMSE), given by:\\
\small
\begin{equation}
M A E=\frac{1}{n} \sum_{i=1}^{n}\left|\hat{y}^{i}-y^{i}\right|
\end{equation}
\begin{equation}
R M S E=\sqrt{\frac{1}{n} \sum_{i=1}^{n}\left(\hat{y}^{i}-y^{i}\right)^{2}} 
\end{equation}
\begin{equation}
M A P E=\frac{1}{n} \sum_{i=1}^{n}\left|\frac{\hat{y}^{i}-y^{i}}{y^{i}}\right| * 100 \%
\end{equation}
\normalsize

\hspace{1cm}where $\hat{y}^{i}$ is the prediction for the $i^{th}$ data point, where the ground truth value was $y^i$. As is evident from the equations, RMSE and MAE are unit dependent while MAPE is a dimensionless quantity. In this survey, while reporting the performance of various regression tasks, we have tried to report the MAPE whenever available. \\

\begin{figure}
    \centering
    \includegraphics[width=5cm]{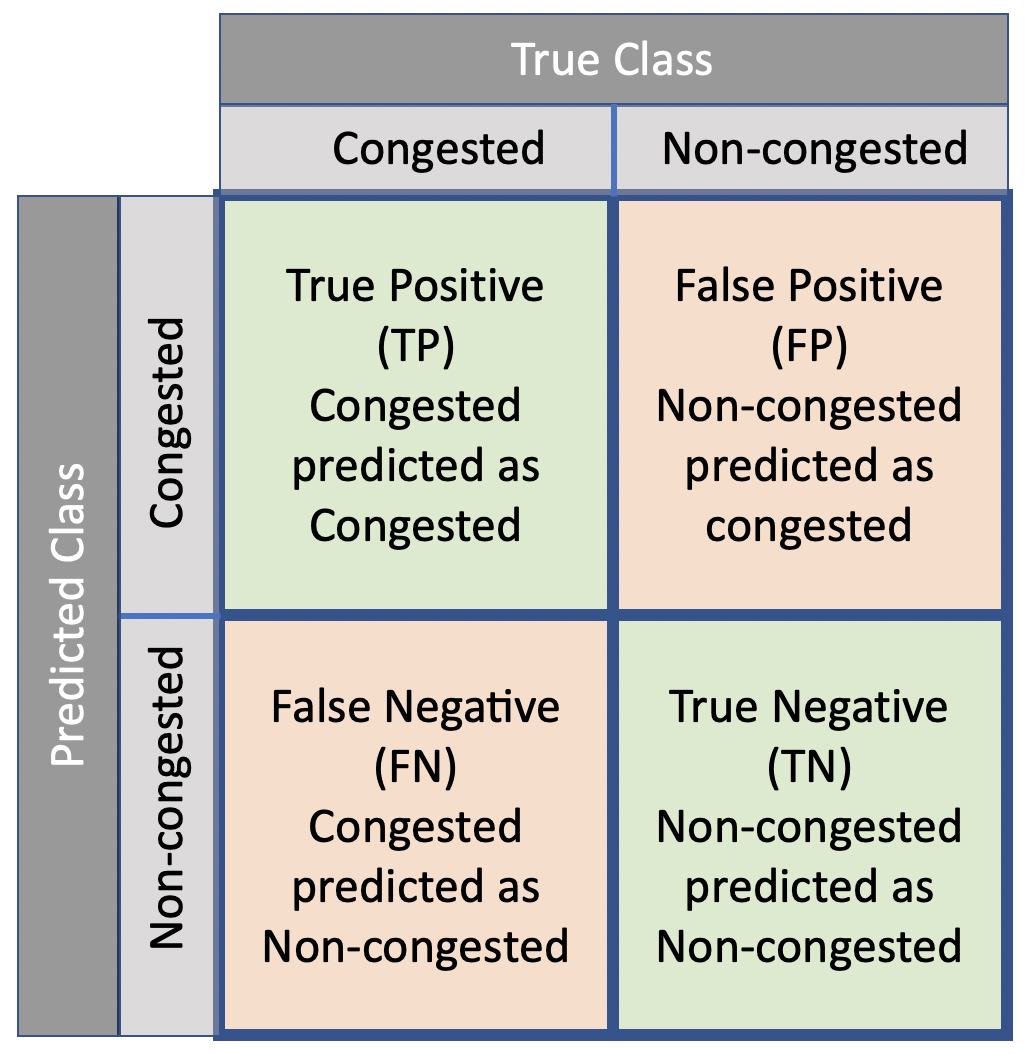}
    \caption{\texttt{Confusion matrix for binary classification task}}
    \label{fig:confusionma}
\end{figure}

The commonly used metrics for classification tasks can be summarised using a confusion matrix as  shown in~\autoref{fig:confusionma}. 
In the light of the confusion matrix, several metrics are defined. The most commonly used metrics are: true positive rate (TPR), true negative rate (TNR) and accuracy, given by:\\
\begin{equation}
    T P R=\frac{TP}{TP+FN}
\end{equation}
\begin{equation}
    T N R=\frac{TN}{TN+FP}
\end{equation}
\begin{equation}
    Accuracy = \frac{TP+TN}{TP+TN+FP+FN}
\end{equation} 

The choice of the metric used to evaluate the performance is often determined by the task at hand. For example, let us consider an example where a deep learning model is being used to classify whether the traffic state is `congested' or `not congested. Let us consider 1440 data points collected every minute over the course of 24 hours and the congestion lasted for an hour and 60 data points have the ground truth label as `congestion'. This implies that even if the model predicts all data points as `not congested', the prediction accuracy is ($\frac{0 + (1440-60)}{1440} = 95.8\%$. Thus, total accuracy is a misleading term for congestion prediction tasks. So, for congestion prediction tasks, the usual practice is to report balanced accuracy (BAC). BAC is  defined as the mean of the TPR for each class ($0.5*(TPR+TNR)$). We observed that some papers presented here refer to BAC as \textit{average accuracy}. Another metric used by one of the papers discussed here is quadratic weighted kappa (QWK~\citep{ben2008comparison}). The QWK metric increases the penalty for classification by chance. The QWK values lie between 0 and 1, with 1 being achieved when the prediction matches the ground truth and 0 when the output is random noise. Mathematically, for an $N$ class classification task, QWK is given by $\kappa=1-\frac{\Sigma_{i, j} \mathbf{w}_{i, j} \mathbf{O}_{i, j}}{\Sigma_{i, j} \mathbf{E}_{i, j} \mathbf{P}_{i, j}} $, where $\mathbf{O_{NxN}}$ is the confusion matrix according to the model prediction and the matrix $\mathbf{E}_{NxN}$  is the expected confusion matrix if prediction was by chance;  $\mathbf{w}_{i, j}=\frac{(i-j)^{2}}{(N-1)^{2}}$ A complete derivation is presented in~\cite{QWKderv}.

The choice of the metric is an important design decision in machine learning tasks and influences the performance of the modelling task. Some common factors which are taken into account while choosing the metric are class imbalance, presence of outliers and invariance properties. The interested reader is referred to \cite{sokolova2009systematic} for a systematic coverage of various metrics for classification tasks.


\subsection{Relevant concepts and terms in transportation} \label{relev3}
In this section, we define some terms from transportation that are fundamental to the understanding of the discussion presented in this survey. A comprehensive review of these definitions and deep insights on their role in traffic prediction is  presented in~\cite{meanspeed1, meanspeed2}, and~\cite[Chapter~2]{gerlough1976traffic}. Here, we selectively reproduce some basic ideas which are necessary for the discussion presented in this survey.
The most common variables used to measure the state of traffic are~\textit{density}, \textit{speed} and \textit{flow}.\\

\textbf{Traffic density:} Traffic density is defined as the number of vehicles per unit length of the road segment. Traditionally, traffic density for the entire road segment has been difficult to measure because of the limited number of sensors to estimate the presence of vehicles. However, this trend is changing with an increasing number of traffic cameras and the advances in computer vision. A related quantity is `occupancy', which is often used as a proxy for measuring density. Occupancy is defined as the percentage of the time during which a point in the road network is occupied by vehicles. Occupancy can be directly measured using sensors, most commonly using Vehicle loop detectors (VLDs). If the length of each vehicle is the same (homogeneous stream of traffic), occupancy is directly proportional to the traffic flow. In practice, the density is most commonly estimated using the fundamental relation between density and speed ($q=k*u$), where $q$ is the flow, $k$ is the density and $u$ is the speed. In the presence of a heterogeneous traffic stream, the relationship between occupancy and density is complex~\citep{heteropgeneity}.\\

\textbf{Traffic speed:} The spot speed (or instantaneous speed) of a vehicle is the speed that is recorded at a given moment in time and at a specified location. This is the speed that is measured on the vehicle speedometer. In transportation engineering, however, we are interested in determining mean speed which can be used as a defining parameter for the traffic stream. In order to compute mean speed, the aggregation can be done in time or space. Space-mean speed, for a given interval of space, is defined as the ratio between the total distance travelled by all vehicles and the total time taken. Time-mean speed, for a given interval of time, is defined as the arithmetic mean of the individual speed of all vehicles. Mathematically, the space-mean speed reduces to the harmonic mean of individual vehicle speeds ($v_i$). Assuming $N$ vehicles,\\
\begin{equation}space~mean~speed = \frac{total~distance~travelled}{total~time~taken} = \frac{N*D}{\sum_{i}^{N}t_i} = \frac{N*D}{\sum_{i}^{N}\frac{D}{u_i}} = \frac{1}{N}\frac{1}{\sum_{i}^{N}\frac{1}{u_i}}\end{equation}
\begin{equation}time~mean~speed = \frac{1}{N}\sum_{i}{N}u_i\end{equation}
The space-mean speed satisfies the fundamental relation between flow, speed and density ($q=k*u$), whereas the time-mean speed does \textit{not} follow the fundamental equation. When using deep learning for predicting traffic speed, the pre-processing steps on the raw data determine which one of them is being predicted. The use of space-mean speed is more common for congestion prediction tasks commonly referred to as the `segment speed. On the other hand, if the data source provides aggregated speed data, the general practice is to predict the same variable~\citep{rathiJi, daganzoKitab}.  \\

\textbf{Traffic flow:} Traffic flow is defined as the number of vehicles passing a reference point per unit time. The reference points are usually chosen in the middle or at the end of a segment.\\

\textbf{Travel time:} Travel time is the time taken by a vehicle to go from point A to point B. Traditionally, travel time has been difficult to measure using aggregated data from point sensors (such as VLDs). With the advent of distributed sensors, such as GPS, these are being increasingly used to estimate the travel time. In the transportation literature, this is referred to as the floating car data (FCD). The challenge however lies in the variation in the percentage of vehicles that share the data at any given point. The major benefit of using FCD is that when the traffic flow is high, more data are collected. This stands in contrast to point sensors where the optimal choice of sensor locations is a major challenge~\citep{FCD}.\\

Two extreme values are important to study the relationship between the aforementioned traffic state variables. Familiarity with these extreme values is necessary to understand the remaining part of this survey and other research dealing with disruptions in road networks. These two values are:
\begin{itemize}
    \item  jam density ($k_j$): The highest possible value of traffic density; this corresponds to traffic speed = 0 km/h.
    \item free-flow speed ($u_f$): The maximum speed at which the vehicles can travel on a given road segment. Under the assumption that drivers respect the speed limit, $u_f$ is the same as the speed limit for the road segment under consideration. 
\end{itemize}

\begin{figure}[h]

\centering
\includegraphics[width=0.7\textwidth]{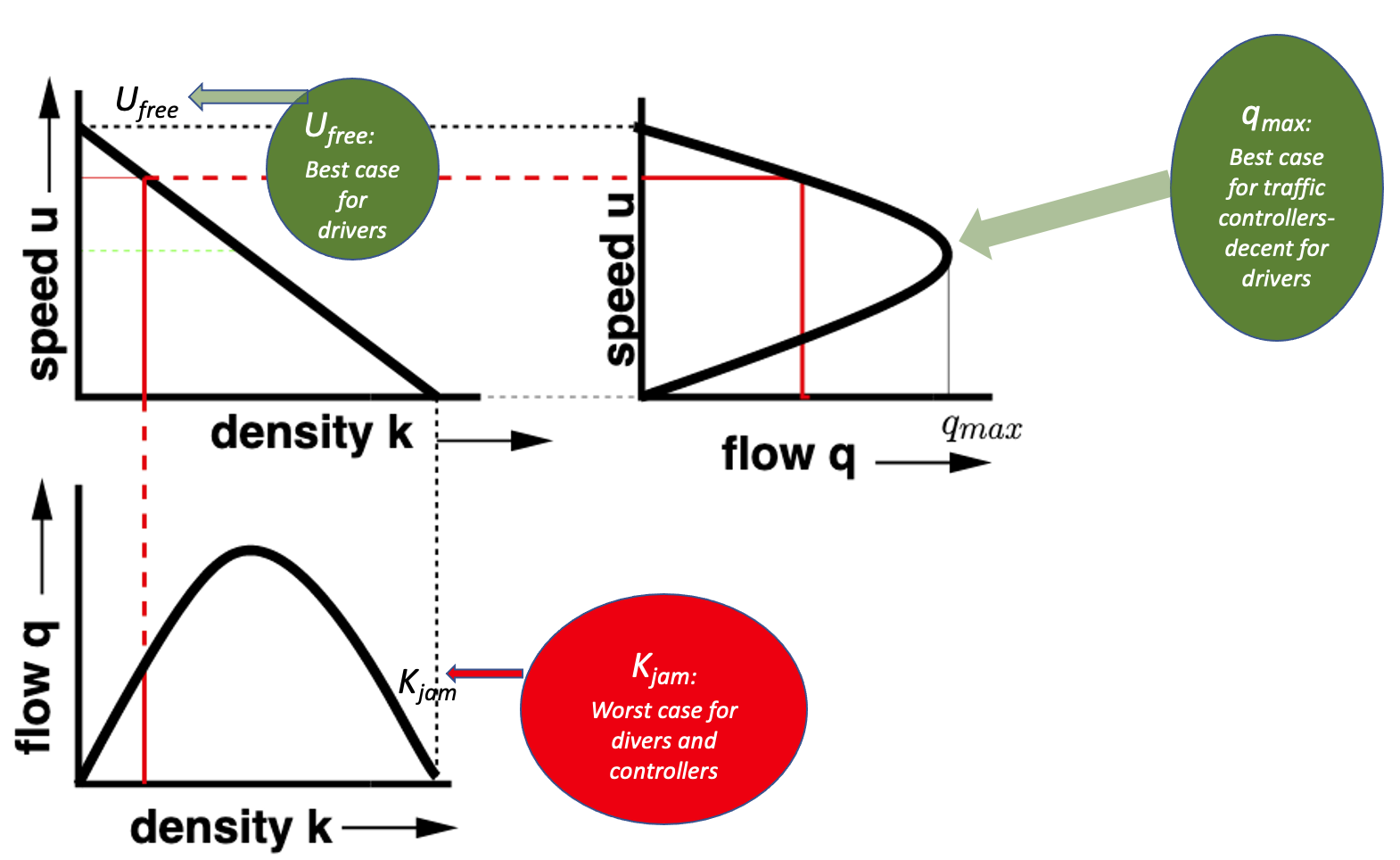}
\captionsetup{margin=2cm}
\caption{\texttt{Representative curves showing the relationship between flow, speed and density. A linear relation between speed and density is assumed here. The two \textcolor{nkGreen}{green} circles point to maximum speed ($U_{max}$) and maximum flow ($q_{max}$) which are the best serviceability conditions w.r.t drivers and traffic controllers respectively. The \textcolor{red}{red} circle points to the maximum density or traffic jam ($K_{jam}$), which implies a lack of serviceability from both perspectives. Base image adapted from}~\href{https://nptel.ac.in/content/storage2/courses/105101087/downloads/Lec-31.pdf}{\textcolor{blue}{NPTEL lecture notes}} creative commons license (\href{https://creativecommons.org/licenses/by-nc-sa/1.0/}{CC-BY-NC-SA})}
\label{fig:fig_flow_density_speed}
\end{figure}

The three variables described above (\textit{speed}, \textit{density} and \textit{flow}) are correlated. However, a generalised equation depicting the relation between these variables has not been established. A simplified linear relationship between speed and density suffices for our discussion here. In~\autoref{fig:fig_flow_density_speed}, we show the relationship between these variables assuming a linear relationship between {speed} and {density}. The various critical points in the three curves in \autoref{fig:fig_flow_density_speed} are highlighted and colour-coded to show the level of serviceability for the two most important stakeholders in a transportation system -- drivers and traffic controllers. 

The choice of the target variable is also motivated by taking into account the consumers of the research output. If the research is targeted at optimising the usage of the transportation network as a system, the focus might be on maximising the throughput of the network; hence the researchers will focus on predicting the traffic flow accurately. On the other hand, if the research is aimed at improving the user travel time, the focus will be on predicting speed or travel time. For instance, when we use a trip planner to find the optimal route from a starting point to a destination, we often want to figure out the fastest route, we are not concerned with the traffic flow on the roads \citep{golledge1995path}. \\

\begin{figure}
\centering
\includegraphics[width=0.4\textwidth]{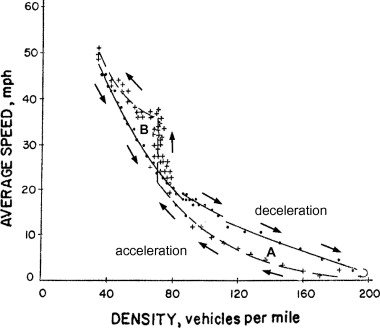}
\captionsetup{margin=2cm}
\caption{\texttt{Hysteresis loop, reproduced from~\citep{hysteresis1975}}}
\label{fig:hysteressi}
\end{figure}

\textbf{Traffic hysteresis:} The fundamental traffic diagram presented in~\autoref{fig:fig_flow_density_speed}, is overly simplified. When real data are plotted using density and speed (space-mean speed), the plot shows significant scatter around an underlying curve~\citep{nicolas}. Researchers have proposed various theories to explain the scatter in the flow-density curve. One such theory is the theory of hysteresis characterised by a distinct loop in the flow-density curve. First observed by~\cite{hysteresis1975}, traffic hysteresis arises due to the human factors in driving. The phenomenon can be easily observed at traffic intersections where a large number of vehicles are queuing. As soon as the signal turns green, the queue does not dissipate at the uniform rate throughout. The queue dissipates from front to back and there are significant time delays. The phenomenon of traffic hysteresis is attributed to the differential acceleration and deceleration rates of vehicles as  shown in~\autoref{fig:hysteressi}. As a result of the hysteresis, a traffic disrupting event continues to affect traffic even after the disruption event has ceased to occur. \\
Another theory used to explain the scatter is capacity drop, attributed to~\cite{cassidy1999some}. They propose that `just before the onset of congestion the outflow out of a bottleneck is higher than in congestion'~\citep[p.451]{van2015genealogy}. The interested reader is referred to~\citep{van2015genealogy} for a detailed review of the traffic flow models. The key takeaways from these efforts  are that the complexity of modelling traffic flow increases as the traffic state moves towards congestion.\\

\textbf{Simulation-based approaches for traffic flow modelling:} When using a microscopic agent-based traffic simulator, individuals and infrastructure elements are modelled as agents. Some examples of traffic simulators are: (1) open-sourced: MATSim \citep{w2016multi}, SUMO \citep{behrisch2011sumo}, SimMobility \citep{adnan2016simmobility}, MATES~\citep{yoshimura2006mates} and TRANSIMS \citep{smith1995transims, nagel2001parallel} and (2) commercially available: AIMSUN~\citep{AIMSUN},~\href{https://www.ptvgroup.com}{VISSIM} and~\href{https://www.paramics.co.uk/en/}{PARAMICS}. In the case of traffic models derived using behavioural methods, a probabilistic model of the possible behaviours (actions and decisions) of each type of agent is programmed into the system by domain experts. The parameters are then calibrated using the available data. During the calibration process, the range of parameter values  is constrained within meaningful ranges of values for each parameter. The initial values and the range of parameters, being set by the domain experts results in the parameters having some level of physical significance, thus making the models {interpretable}. The predicted traffic is the net result of the interaction between the agents in the calibrated model. \\

\textbf{Short-term traffic prediction:} Some model-based approaches to short-term traffic prediction are  DynaMIT~\citep{dynamit1} and DYNASMART~\citep{mahmassani2004dynasmart}. In model-based approaches, the central algorithm is Dynamic Traffic Assignment (DTA)~\citep{janson1991dynamic}. 



\subsection{Definition and classification of congestion}\label{defining-congestion}

`Congestion can be defined as the phenomenon that arises when  the  input  volume  exceeds  the  output  capacity  of  a \textit{facility}' \citep[Section 2.1]{stopher2004reducing}. Depending on the number and size of \textit{facilities}, congestion results in varying levels of loss in the serviceability of the road network. The literature on congestion prediction defines congestion either in terms of one of the traffic state variables ({speed}, {density},  {flow}) or in terms of derived variables such as the ratio of average speed to speed limit. A recent survey of the traffic variables used to define congestion is presented in~\cite{afrin2020survey}. Once a variable is chosen, the values of the variable are quantified into a fixed number of levels in order to define a classification task. A binary quantisation can be achieved by using a single threshold on any one of the traffic variables. For example, a traffic density of more than a certain threshold can be referred to as `{jam}' and vice-versa.

Based on the spatio-temporal frequency of their occurrences, congestion can be classified into two types- recurring and non-recurring. Recurring congestion, as the name suggests, is the congestion that manifests itself repeatedly in space and/or time. Specific areas  of the city might experience traffic jams regularly at specific times during the day or on certain days of the week. non-recurring congestion does not follow a spatio-temporal pattern.~\cite{mcgroarty2010recurring} present a summary of the causes behind both types of congestion. They report that recurring congestion is almost always caused by an infrastructural bottleneck. On the other hand, the non-recurring congestion can be caused by unforeseen events such as extreme weather conditions, natural and man-made disasters, accidents or planned events such as big concerts and roadworks. In the process of reviewing the literature, we observed that a significant number of papers have not clarified whether they attempted to predict recurring or non-recurring congestion. We have carefully evaluated their results and included them in their respective sections.

\subsection{Synergies between model-based and deep learning based traffic prediction}\label{two_pillars_section}
The data-driven methods for short-term traffic prediction on the other hand do not take the user behaviour into consideration. Data-driven models aim to solve the prediction task by assuming traffic to be a measurable state of the system and attempt to predict its state into the future. Specifically, the workflow is to use all available data from sensors and output the predicted traffic state variable. Particularly, when deep learning models are used for this task, the model internals (weights) have no physical significance. Due to the lack of {interpretability}, extensive validation is necessary to ensure that the deep learning model predictions are useful.\\ 

   Traffic simulators are useful for investigating the effects of new policies. Their importance further increases in studies where real data are unavailable or cannot be collected. For example, a simulator can be used to study the effects of a city-wide failure of traffic lights. Real-world data cannot be obtained at such a scale; therefore the researchers rely on the fact that the behaviour-modelling of drivers and the modelled physical interactions between agents, together provide reliable inferences. \\
   
   Synergies between model-based and data-driven approaches can benefit the research using both approaches. Congestion and accident databases suffer from severe class imbalance problems.~\cite{fukuda2020short} used a traffic simulator to produce traffic data after simulated accidents. The generated data was then used for training a deep neural network. As we shall discuss in~\autoref{TSCsection}, when using a deep reinforcement learning framework to determine the optimal network-level control measures for congestion alleviation, a microscopic traffic simulator is incorporated in the framework.~\cite{borysov2019generate} used a deep generative model to generate agents for a simulation platform.\\
   
   Deep learning models are increasingly being used to \textit{learn} the physics behind the nonlinear dynamics of complex networks. Such models have typically been studied under the heading  of Physics informed deep learning (PIDL)). PIDL was conceptualised with two motivations. First, PIDL enables us to use prior domain knowledge to regularise the function being approximated by the deep learning model, thereby reducing overfitting~\citep{raissi2017physicspart1}. Second, PIDL can be used to discover new partial differential equations from the data~\citep{raissi2017physicspar2}. When PIDL models are used in the traffic domain, microscopic traffic simulators often play an important role in the training process of such models. By design, the traffic simulators respect the traffic flow dynamics and hence can be used to regularise the traffic state predictions from a neural network. For example, SUMO was used for this purpose in~\cite{PDIL-SUMO}. The PIDL models hold a lot of potential because several drawbacks of deep learning models can be addressed. The PIDL models are more robust to missing data, noise, overfitting and might help in making deep learning models interpretable. However, the PIDL research in transportation is still at a nascent stage with very few papers and has not been covered in this survey. The interested reader is referred to~\cite{Shietal2021a}, who used data from the Next Generation SIMulation~\href{https://ops.fhwa.dot.gov/trafficanalysistools/ngsim.htm}{(NGSIM)} dataset from the US Department of Transportation and proposed a PIDL model with two components, one data-driven and the other, model-driven. The influence of each component in the training process can be controlled using a single parameter. In related work,~\cite{Shietal2021b} used a hybrid PIDL model on the same dataset to estimate the parameters of the second order partial differential equations governing traffic flow. Other recent efforts using PIDL are:~\cite{PDIL3}, who present a detailed comparison between DL and PIDL models and report that PIDL models are faster to train ($>$50\% faster) and perform better than other DL models when the sensor locations are fixed,~\cite{PDILturbule} who proposed a PIDL model called~\href{https://github.com/Rose-STL-Lab/Turbulent-Flow-Net}{Turbulent-Flow Net} for predicting turbulent traffic flow and~\cite{PDIlcarfollow}, who used a PIDL model to learn the dynamics of car-following models. To summarise, these efforts demonstrate promising avenues for further research into the synergies between model-based and data-driven techniques.

\section{Previous surveys and organisation of this survey}
The nonlinear activation functions in deep neural networks can capture the nonlinearities in the traffic data \citep{polson2017deep}. As discussed in~\autoref{section:machineTrans}, the {depth} of the network enables us to model high-level features in the data. Traffic data are characterised by variations over space and time. Two specialised neural network architectures, CNN and RNN  (also discussed in \ref{section:machineTrans}) are very helpful in capturing these variations. CNNs are useful while modelling spatial inter-dependencies whereas RNNs are useful while modelling temporal variations in the data. During the course of our literature review, we found that most of the successful neural network architectures for traffic prediction were designed using CNN and RNN units as building blocks.\\

 An extensive summary of deep neural networks in various aspects of transportation systems is presented in ~\cite{WANG2019144}. They cover a wide range of traffic related prediction tasks using deep neural networks -- traffic signal identification, traffic variables prediction, congestion identification and  traffic signal control.~\cite{nguyen2018deep} also cover the aforementioned tasks and add three other tasks to the list,\textit{ viz.} travel demand prediction, traffic incident prediction and driver behavior prediction.~\cite{wang2020deep} surveyed the applications of deep learning in various domains which use spatio-temporal data (transportation, human mobility, crime analysis, neuroscience and location-based social networks). Their review comprises recent papers dealing with deep learning methods for tasks such as traffic variables prediction, trajectory classification, trajectory prediction, and travel mode inference.~\cite{9046288} present a taxonomic survey of Graph Neural Networks (GNN) and highlight the applications of GNNs in different fields, including transportation.~\cite{XIE20201} summarise various approaches where deep learning was used for the most common types of flows in a city -- crowd flows, bike flows, and traffic flows.\\

\begin{figure}[h]
    \centering
    \includegraphics[width=0.95\textwidth]{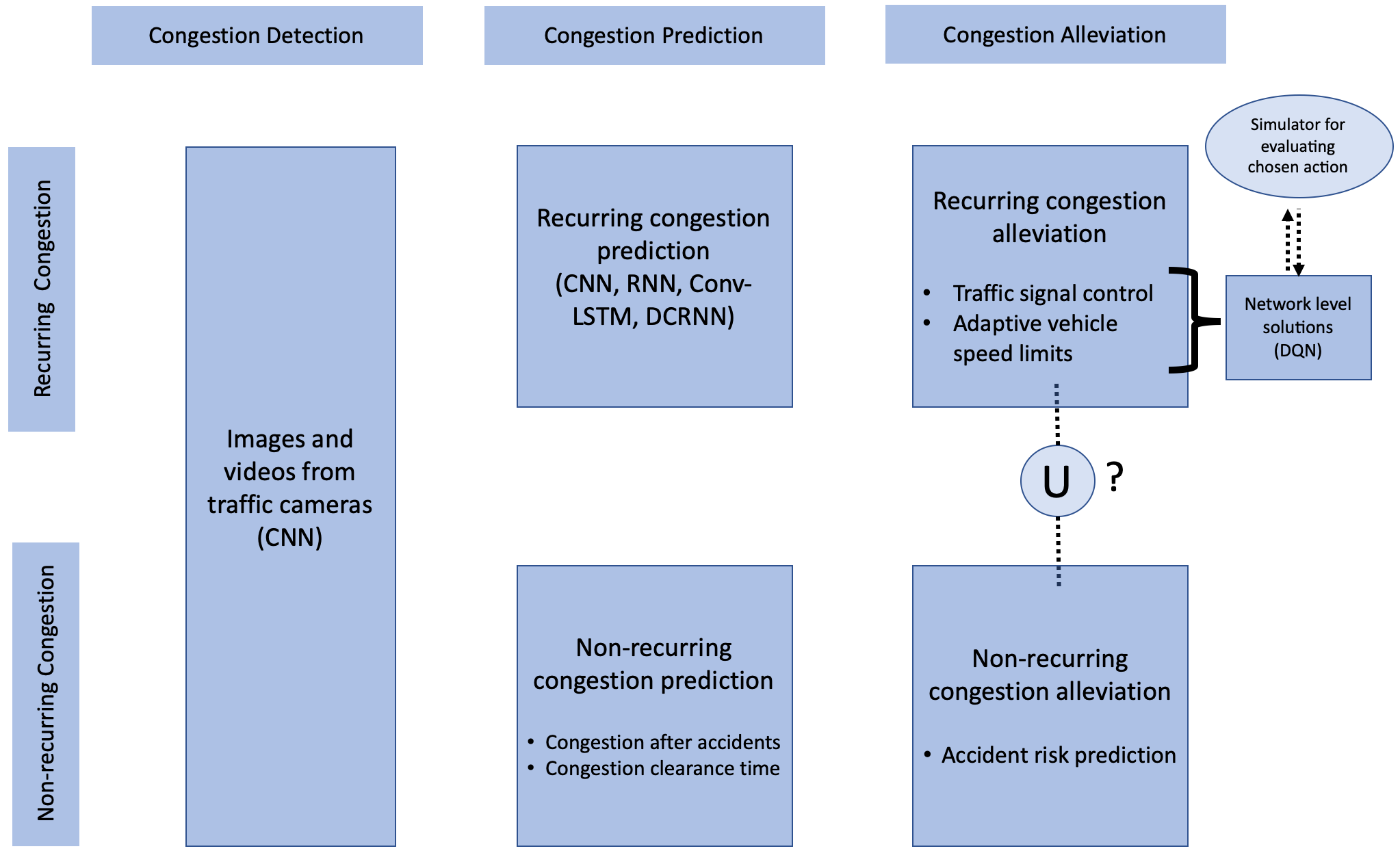}
    \captionsetup{margin=2cm}
  \caption{\texttt{Overall flowchart showing various sections of the survey. The question mark denotes the potential link between non-recurring and recurring congestion alleviation}} 
    \label{oberallf}
\end{figure}

Congestion prediction refers to the prediction of traffic state variables when congestion is imminent. It is a special case of traffic prediction. The relative difficulty of congestion prediction tasks in comparison to traffic prediction can be attributed to the instability of the traffic dynamics beyond the point of maximum flow~\citep{chung2011quantification}. This higher relative difficulty is also obvious from the fact that the performance of a typical traffic prediction model degrades as the state of traffic approaches. The importance of deep learning for congestion prediction is also due to the relatively higher stability of deep learning models compared to other data-driven approaches \citep{yu2017deep}. To the best of our knowledge, a comprehensive survey paper covering the applications of deep learning in congestion-related prediction tasks does not exist. This survey paper is an attempt to bridge this gap in the literature. We discuss the applications of deep learning in the detection, prediction and alleviation of both types of congestion
-- recurring and non-recurring. The section on congestion prediction is not differentiated into recurring and non-recurring cases because the deep learning models which detect congestion from traffic images, detect both types of congestion. Whenever possible, we have attempted to incorporate the challenges from a policymaker's point of view. The inclusion of the policymaker's outlook is important in order to make research outputs deployable in a real world setting. The scope of the current survey is presented in~\autoref{oberallf}.\\

The key design aspects of the deep learning architecture and the key aspects of the dataset used in each paper have been summarised so the reader can refer to such papers when working on similar datasets. Some papers present an extensive sensitivity analysis of their models. We have reproduced and highlighted the crucial insights (if any) in the summary presented at the end of each subsection. \\
Most research papers covered in this survey were published in the period 2016-2021. Sometimes, in order to briefly discuss the background of some algorithm paradigms, we have referred to classical papers from the past.

\section{Deep learning for congestion detection}
With increasing access to new data sources, new opportunities to automatically detect traffic congestion are being explored. Unlike the other two sections, the congestion detection models are not differentiated for recurring and non-recurring congestion. This is so because congestion detection models invariably detect both types of congestion irrespective of the causal factor behind them. They are instead differentiated on the basis of the data source used. The most commonly used sources to detect both types of congestion are images and videos obtained from traffic cameras.\\


The benefit of using traffic camera images is that without exception, all vehicles are captured in the traffic image. Thus, other factors like penetration ratio (percentage of vehicles being tracked) do not need to be considered. With the increasing number of cameras on roads, the cognitive load on the human identifying congestion from the images is high. In order to reduce the cognitive load, deep learning has been widely applied for the automatic detection of congestion from traffic images. Deep learning models which are known to perform well for Computer Vision (CV) tasks have been adopted to detect traffic congestion. CV refers to the task of extracting useful information from images.\\ 

Convolutional Neural Networks (CNNs) form the building block of commonly used  deep learning architectures for image classification tasks. The seminal works in this field were: AlexNet \citep{krizhevsky2012imagenet}, InceptionNet \citep{googlenet}, Resnet \citep{resnetpaper}, R-CNN \citep{girshick2014rich}, Mask-RCNN~\citep{maskrcnnpaper}, VGGNet~\citep{vggnet}, and YOLO \citep{yolopaper}. Deep neural networks pre-trained on large image datasets like ImageNet~\citep{imagenetData} and COCO~\citep{COCO-data} are readily available. Three approaches are possible when using well-known architectures for traffic image classification. When the number of traffic images in the dataset available at hand is very high (order of 10000 images), these models can be trained from scratch using the available data. When the dataset available is small, the weights for the adopted deep learning  model are initialised to the available pre-trained model. The third approach is to retain the pre-trained model as it is and add another module in sequence. When applied to images, the deep learning models can be used to estimate the total number of vehicles in an image, thereby allowing us to estimate the traffic density. When applied to video data (a sequence of images), the deep learning models can be used to estimate traffic speed. \\

Two variations of CNN based architectures (AlexNet and YOLO) are utilised \citep{chakraborty2018traffic} to detect congestion using a binary classification into traffic images collected from 121 cameras from Iowa, USA over  a period of 6 months. Manual labelling of traffic images into congestion and non-congestion labels is a time-consuming task. The authors, therefore  use occupancy data obtained using vehicle loop detectors (VLDs) to automatically label the images into two classes based on occupancy ($\>20\%$ occupancy is labelled as `congested'). The reported accuracy for detecting congestion was 90.5\% for AlexNet and 91.2\% for YOLO respectively.~\cite{8397881} compare two variations of AlexNet and VGGNet to detect congestion on traffic images obtained from more than 100 cameras from Shaanxi province, China. Their dataset is highly varied-~comprising images for day and night traffic and varying weather conditions. Their results show comparable performance for both architectures (78\% for AlexNet compared to 81\% for VGGNet). They report that AlexNet is significantly faster to train due to the smaller size of the neural network. They use binary classification (`{jam}' or `{no jam}').~\cite{impedovo2019vehicular} compared the performance of YOLO and~\href{https://paperswithcode.com/media/methods/Screen_Shot_2020-05-23_at_7.44.34_PM.png}{Mask-RCNN} on three manually labelled datasets obtained from two traffic image data sources (\href{http://agamenon.tsc.uah.es/Personales/rlopez/data/rtm/}{GRAM} and \href{http://visal.cs.cityu.edu.hk/downloads/trafficdb/}{Trafficdb}). The three datasets are of varying image quality- first, comprising 23435 images at low resolution (480x320p), second, comprising 7520 frames at mid resolution (640x480p), and third comprising 9390 frames at high resolution (1280x720p). They achieve congestion detection in two steps. The first step focuses on identifying the number of vehicles in each frame. In this step, the Mask-RCNN achieves an accuracy of  46\%, 89\%, 91\% respectively while YOLO achieves an accuracy of 82\%, 86\%, 91\% respectively. The performance of YOLO is resistant to the image quality and the training time is almost half of Mask-RCNN. They select YOLO as the object detector model and use its output as the input to the second step. In the second step, they use Resnet on the output of YOLO to predict traffic congestion as a multiclass classification task (3 classes). The reported accuracies for light, medium and heavy congestion are 99.7\%, 97.2\% and 95.9\%. \\

A CNN model is used in~\cite{KURNIAWAN2018291} in order to classify traffic images obtained in Jakarta, Indonesia. The data are collected for 15 days and 14 camera locations are used. They use manual labelling of the traffic images into `jammed' and `not jammed' classes. The reported average accuracy using 10-fold cross-validation is 89.5\%.~\cite{rashmi2020vehicle} investigated the performance of YOLO when the traffic is highly heterogeneous. They collect one week of data from Karnataka India. They use transfer learning with a YOLO model pre-trained on the COCO dataset. While counting vehicles in the images, YOLO performs well (accuracy between 92\% and 99\%) for buses, cars and motorcycles but when predicting the modes of transport which are specific to the zone of study, the performance drops below any useful level.\\

\textbf{Summary:} We observe significant differences in the image quality based on the data source. This results in differences in the model performance. The traffic images obtained from developing countries present a major challenge due to the large heterogeneity of the traffic stream. Another major difference between datasets is that when alternate sources of data are present (such as VLDs), the labels for training data can be created automatically, instead of manual labelling. If the quality of the images obtained from traffic cameras is not high, deep learning based image super-resolution can be used for improvement. Image super-resolution refers to the task of increasing the resolution of input images. Deep learning based super-resolution has been widely researched in the computer vision community, but we are yet to see its applications in improving the quality of traffic images. A comprehensive survey on deep learning applications for image super-resolution is presented in~\cite{super-resolution-survey}. Another potential avenue for improvement in congestion detection from cameras is deep learning based video frame rate increment~\citep{archer-supervisor}.




\section{Deep learning for congestion prediction}\label{section:Deep learning for congestion predc}
\subsection{Deep learning for recurring congestion prediction}\label{subsdlrc}
Recurring congestion occurs due to infrastructural bottlenecks, which are insufficient to handle the peak demand. By definition, recurring congestion occurs at familiar locations in the network. In the light of this definition, the specific task while predicting recurring congestion is the prediction of daily variations in the time of occurrence and the severity of recurring congestion. The most commonly used congestion prediction task is binary classification (`jam' or `no jam'). Some papers predict congestion as a multiclass classification task (`light', `medium' and `heavy' congestion). It should be noted that some papers included in this section did not predict traffic congestion but instead focused on short-term traffic prediction as a regression task (predicting speed, density, flow, queue length etc.). Such papers have been included if their model performance was reported to be stable when the state of traffic approaches congestion. For each paper listed here, we have reported the key aspects of their performance metric and the key takeaways from their sensitivity analyses. The papers discussed in this subsection are summarised in~\autoref{tab:predict_Recurrent_congestion}.\\

\textbf{Using LSTM models}: An LSTM model is used in~\cite{yu2017deep} to predict traffic speed during peak hours. They use vehicle loop detector (VLD) speed recordings from the publicly available \href{https://dot.ca.gov/programs/traffic-operations/mpr/pems-source}{Caltrans} dataset from California, USA and attempt to predict the next hour traffic speed at each sensor location. When predicting the traffic speed during peak hours, their LSTM model is reported to achieve a MAPE of 5\%. An important observation from their study is that timestamp features (encoded values for the time of day and day of the week) significantly improve the prediction performance during the peak hour. This practice of incorporating timestamps has been adopted in papers that appeared later. In this paper, each VLD is modelled separately. In the subsequent papers, researchers have developed techniques to incorporate spatial information into the LSTM models and hence propose a single model for predicting congestion at several intersections at the same time.~\cite{rahman2020real} use an LSTM model to predict the queue lengths at intersections by incorporating the spatial information in an efficient manner. In order to predict the queue length at a query intersection for the next traffic cycle (red signal), their model takes as input the queue lengths of the query intersections and two upstream intersections at the current cycle. They then attempt to predict the queue length at the query intersection for the next cycle. They use VLD data collected at 11 intersections for a period of three months in Orlando, Florida. When predicting queue lengths, they report an average RMSE  close to one (as inferred from their plot).\\

\textbf{Comparing LSTM and CNN:} A comparison of the performances of convolutional neural networks (CNNs) and recurrent neural networks (RNNs) for predicting traffic congestion is presented in~\cite{sun2019traffic}. Their dataset comprises 28 days of GPS-trajectories of 2000 taxis in Chengdu, China. They used map matching to map the GPS trajectories to road segments and calculate average speed during 5-minute time slots for each segment. They then use the average speed values to define four congestion \textit{levels} based on the average traffic speed. The reported RMSE for the average speed prediction for their best models is 3.96 \textit{km/h}. The classification accuracy is then reported for the predicted level of congestion. They conclude that, given a sufficiently long input horizon (90 minutes), the performance of CNN models is as good as recurrent network models. We believe this is an important observation for two reasons. First, since the CNN models are faster to train because by construction they support GPU parallelisation in a more efficient way, it can save time on the part of researchers if CNNs are explored as a modelling option before exploring LSTMs. Second, since LSTMs are typically used to capture long-term dependencies in traffic data when used independently, they might not be very useful for short-term traffic prediction. The following papers in this subsection use combinations of CNNs and RNNs to design specialised architectures.\\

\textbf{Using a combination of LSTM and CNN:}~\cite{liu2017short} present~\href{https://ieeexplore.ieee.org/mediastore_new/IEEE/content/media/8123608/8170872/8171119/8171119-fig-3-source-small.gif}{Conv-LSTM}, which is composed of CNN and LSTM units. The convolution operations are used to capture the spatial dependencies. The output of convolution operations is used as inputs to the LSTM units. While using their Conv-LSTM model to predict traffic flow, they achieve a MAPE of 9.53\%. Additionally, they include a bi-directional LSTM module to include the effects of historical data and achieve a lower MAPE of 6.98\%. The performance is reported to be stable across varying levels of traffic flow, hence we have included their paper as a congestion prediction model.~\cite{ranjan2020city} predict city wide traffic heat maps for three prediction horizons of 10, 30, and 60 minutes. Since the input and output heat maps have the same dimensions, they propose a symmetric U-shaped architecture with CNN blocks at both ends (inputs and output ends). The bottleneck layer (at the highest depth) is made up of four LSTM units and skip connections are used to connect the CNN outputs at various depths. The proposed architecture is called~\href{https://www.researchgate.net/profile/Navin-Ranjan/publication/341062590/figure/fig4/AS:895246355398656@1590454772095/PredNet-model-architecture-a-The-model-is-divided-into-three-sections-section-A-is.png}{PredNet}. The dataset consists of traffic heat maps based on space mean traffic speed collected from Seoul, S. Korea. The temporal resolution of their data is 5 minutes. Congestion is defined as a three-level variable based on the average speed of traffic. They report the performance of PredNet to be stable with the increasing lengths of output horizon. When predicting congestion 60 minutes into the future, PredNet achieves a mean accuracy of 84.2\% (compared to 75.67\% when Conv-LSTM is used on the same data). PredNet is also reported to be significantly faster to train ({\raise.17ex\hbox{$\scriptstyle\sim$}}8 times faster) compared to the Conv-LSTM model discussed previously in this paragraph.\\

\textbf{Special focus on the heterogeneity of the road network}:
Some papers have highlighted the differences in the complexity of the congestion prediction task based on the heterogeneity of the road network. Such observations are particularly reported when the congestion prediction is attempted at large parts of the network for multiple time steps.~\cite{shin2020prediction} use a three-layered LSTM network to predict the congestion levels in data collected from urban and suburban areas in and around Seoul, South Korea. The total number of road links was 1630 and the data were collected for one month at a resolution of 5 minutes. The dataset had 33\% of the records missing. In order to handle the missing data, they propose a trend-filtering based spatio-temporal outlier detection and data correction algorithm. The model predicts traffic speeds but outputs congestion levels based on thresholds recommended by the local policy-making authority. The model performance is stable across the entire range of traffic speeds, hence we have included their work as a congestion prediction model. They report differences in the performance of the model when predicting traffic speed for two different types of road networks around Seoul (MAPE of 4.297\% for suburban vs MAPE of 6.087\% for urban roads). The mean absolute error (MAE) however, was higher for  some suburban roads compared to urban roads (urban: 2.54 km/h, suburban: 2.78km/h). They acknowledge that high MAE error for suburban roads, is in fact, misleading because the average speeds for suburban roads are higher. The takeaway from their paper is that while applying deep learning for predicting congestion, different types of roads present different complexity.\\
~\cite{cheng2018deeptransport} propose a specialised~\href{https://d3i71xaburhd42.cloudfront.net/56f17fa9a094e80f3e5e47311f2f31613d48e7b0/4-Figure4-1.png}{architecture} built using CNNs, RNNs and an attention mechanism to predict congestion levels. Their dataset, called \href{https://github.com/cxysteven/MapBJ}{MapBJ} consists of 4 months of data collected at 349 road links in Beijing at a temporal resolution of 5 minutes. Each road link is labelled into one of the four congestion levels (fluency, slow, congestion, extreme congestion), based on a speed-limit normalised variable called `limit level'. The exact mathematical representation of the `limit level' is not presented, however, the idea is similar to using a ratio of the actual speed to the speed limit, given by ($\frac{average~speed}{speed~limit}$). The road network is converted to a representative graph with road links being represented as vertices and intersections as edges. The upstream and downstream vertices of the target vertex are grouped by vertex order. The input traffic conditions from different groups of vertices are used as inputs to the CNN module, followed by the RNN module. Thereafter, an attention model is used to assign different weights to different groups of vertices. The separate modules for upstream and downstream roads allow insights into the effects of upstream and downstream links on the congestion prediction at the target link. While predicting congestion state 60 minutes into the future, \textit{higher} weights are observed for higher-order downstream neighbours. On the other hand, \textit{lower} weights are observed for higher-order upstream neighbours. Such analyses provide useful insights into the demand and the flow of traffic. They use a metric called \textit{quadratic weighted Kappa (QWK)}~\citep{ben2008comparison}. They report an average {QWK} of around 0.6. They report QWK values to be  0.69, 0.63, 0.57 and 0.52 for predicting at 15, 30, 45 and 60 minutes respectively. For comparison, the QWK for a stacked autoencoder model was 0.68, 0.62, 0.56 and 0.49 for predicting at 15, 30, 45 and 60 minutes respectively.\\

\textbf{Large-scale recurring congestion prediction (Congestion propagation)} Congestion propagation can be understood as a special case of congestion prediction. It is the study of the evolution of congestion in a larger part of the network than what is usually covered by congestion prediction models. Congestion propagation can be studied under the same heading as congestion prediction. However, during the literature search for this survey, we observed that special challenges are encountered when deep learning models are used to predict traffic congestion for the entire network.\\
A specialised architecture called \href{https://i.loli.net/2019/04/23/5cbec05ce557e.png}{DCRNN} was proposed in~\cite{DCRNNWALAPAPER} in order to predict traffic flow for several time steps (15 minutes, 30 minutes and 1 hour). The dataset used is~\href{https://dot.ca.gov/programs/traffic-operations/mpr/pems-source}{PeMS}. The  model performance is unchanged during peak hours and under varying levels of flow, hence this paper has been included as a congestion prediction model. DCRNN consists of an encoder and a decoder component. The encoder takes traffic flow data with spatial parameters encoded into a graph and outputs \textit{hidden states}. The decoder attempts to predict the next-step traffic flow, either using the \textit{hidden states} from the encoder with a probability $\rho$, or using the ground truth data with a probability ($1-\rho$). At the start of the training, the value of $\rho$ is close to 1 and is decreased to 0 by the end of the training. The reported MAPE are 2.9\%, 3.9\% and 4.9\% for 15, 30 and 60 minutes respectively.~\cite{andreoletti2019network} also used a DCRNN model to predict congestion as a binary classification task based on traffic density. The threshold for binary classification is defined using road link specific load factors ($\alpha$). The load factor ($\alpha$) for a link is defined as the ratio of current traffic density to that of the average traffic density ($\alpha=\frac{{current}~{traffic}~{density}}{{average}~{traffic}~{density}}$). The reported congestion prediction accuracy is 96.67\% when $\alpha$ is set to 3. At high values of $\alpha$, the rate of false negatives (FN) increases. A high value of alpha implies that only very high traffic density is classified as congestion. A value of $\alpha=3$ appears to be optimum with a false negative rate of 2.4\%. The optimal choice of threshold in order to achieve better generalisation of deep learning models has been explored further in papers that appeared later on.\\
\cite{ma2015large} used a Restricted Boltzmann Machine (RBM) combined with a recurrent neural network (RNN) model to predict the evolution of congestion. The dataset comprises GPS-trajectories of taxis plying on 515 road links in Ningbo, China. They use a \textit{network-wide} threshold on traffic speed to determine whether the predicted traffic speed implies congestion. They report an average accuracy of 88.2\%. An interesting observation from their sensitivity analysis is that an increase in the threshold degrades the model performance. They hypothesise that this might be due to the higher fluctuations in the congestion propagation patterns when a higher percentage of road links fall into the congested category.~\cite{7966128} used the~\href{https://dot.ca.gov/programs/traffic-operations/mpr/pems-source}{PeMS} dataset from California and proposed a distributed network where each intersection of the road network was modelled using a separate deep learning model. They used a combination of CNN and RNN architectures to predict the congestion levels. In order to binarise congestion, they introduced \textit{node-specific} thresholds, instead of a \textit{network-wide} universal threshold, thereby giving their model more expressive power compared to~\citep{ma2015large}. The node-specific thresholds are defined using the ratio of the average speed to the speed limit (${threshold}=\frac{{average~speed}}{{speed~limit}}$). The sensitivity analysis of their model revealed a \textit{drop} in the model's performance as the congestion levels increase. They hypothesise that this {drop} is due to class imbalance (high number of data points for non-congested cases). They attempted to remedy this \textit{drop} by preferentially weighing the data points where congestion was high. The preferential weighting is achieved by modifying their mean squared error loss function to include a penalty ($w_i*\beta_i$) for each data point $i$. The difference between the predicted flow and the ground truth is  $\beta_i$ . The value of $w_i$ is equal to 1 if the \textit{traffic flow} for data point $i$ is less  than $0.5*(max~flow) $ and $0$ {otherwise}. The  $\beta$ value is a measure of the prediction error. Now, using the variable $w_i$, the model imposes an extra penalty for errors when the {traffic flow} was more than half the maximum flow. \\
\cite{wang2016traffic} proposed~\href{https://ieeexplore.ieee.org/mediastore_new/IEEE/content/media/7837023/7837813/7837874/7837874-fig-1-source-large.gif}{erRCNN}, which is built using CNN units followed by RNN units. In~\autoref{fig:hysteressi}, we observed that when the flow exceeds beyond a threshold, the speed-density curve becomes scattered and abrupt changes in the average speed are observed. Their erRCNN architecture is shown to handle these abrupt changes. The error-correcting RNN allows the model to be updated when the prediction performance drops due to a change in the state of traffic. Thus, the model is capable of handling streaming data. They used a GPS dataset collected from 2 major ring roads in Beijing, China. The reported RMSE for speed prediction varies between 5 km/h when the prediction horizon is 10 minutes and 8 km/h when the prediction horizon is 50 minutes. Additionally, in order to understand the sources of congestion propagation, they propose a metric called segment importance. Each segment influences the traffic on other segments. If the road in question has $n$ segments, it can be assumed that the trained errRCNN model has learnt  a mapping between traffic speeds at each segment at time $t$ and the traffic speeds at each segment at time $t+1$. Mathematically, ${erRCNN}~([([v^{t}_{seg_1}, v^{t}_{seg_2}.. v^{t}_{seg_n}])]) = [v^{t+1}_{seg_1}, v^{t+1}_{seg_2}.. v^{t+1}_{seg_n}]$. Using this assumption, first they define the influence of $seg_i$ on $seg_j$ as the derivative of $v^{t+1}_{seg_j}$ w.r.t $v^{t}_{seg_i}$. Finally, they define the segment importance of a segment as the sum of the influences to all other segments. They then map the segments with high segment importance to physical locations in the network and uncover some sources of congestion (such as an intersection connecting two highways). Even though these observations are intuitive, their contribution is significant because it throws light on how deep learning models can be used to provide interpretable insights, thereby promoting wider acceptance by transport authorities. \\

\textbf{Deep learning for congestion trees:} Another popular method that has been historically used to model the propagation of congestion utilizes the concept of congestion trees. Attempts have been made to model the evolution of congestion trees using deep learning. A congestion tree is formed when congestion on one road segment results in the building up of congestion onto the adjacent road segment. Several congestion trees can be combined by removing the redundancy between them. This gives rise to a congestion graph.~\cite{di2019traffic} remove the redundancy between the congestion trees by creating a directed acyclic graph  (DAG) through combination of the congestion trees. This DAG is then converted into a spatial matrix of congestion levels, with each cell of the matrix representing at most one segment. The spatial matrix helps preserve the adjacency information between the road segments. A sequence of these spatial matrices (SM) is then passed to a~\href{https://ieeexplore.ieee.org/mediastore_new/IEEE/content/media/8778425/8788716/8788784/336300a298-fig-4-source-large.gif}{Conv-LSTM} model for predicting the SM at the next time step. The predicted SM is transformed back to the congestion graph and then to the congestion tree, which can then be used to provide a visual representation of the predicted evolution of congestion. When using a 5-minute prediction horizon, they reported the Mean Squared Error (MSE) of 0.27 for weekdays and  0.07 for weekend traffic. MSE for 15 minutes was 0.73 (weekday) and 0.37 (weekend). In their result, MSE has no units because it is not computed for a traffic variable, but for the spatial matrices. For comparison, the LSTM model achieved an MSE of 0.59 for weekdays and 0.32 for weekends when using a 5-minute prediction horizon.\\

\begin{table}[]
\begin{adjustwidth}{-2cm}{}
     \caption{\normalfont{Table summarising the papers discussed in~\autoref{subsdlrc}} ({Deep learning for recurring congestion prediction })}
    \label{tab:predict_Recurrent_congestion}
    
    \begin{tabular}{l l l l l l}
    
    \toprule
   Paper 
  &\shortstack[l]{Congestion \\defined on\\the basis of: }
  &\shortstack[l]{DNN \\ architecture }  
  &\shortstack[l]{ Performance} &\shortstack[l]{Data\\source} 
  &\shortstack[l]{Unique\\aspect}  \\
          \toprule
 \shortstack[l]{\citep{wang2016traffic} }
  &\shortstack[l]{Traffic speed}
  &\shortstack[l]{~\href{https://ieeexplore.ieee.org/mediastore_new/IEEE/content/media/7837023/7837813/7837874/7837874-fig-1-source-large.gif}{erRCNN}\\(built using\\ CNN, RNN)
} 
  &\shortstack[l]{5km/h$\leq$RMSE$\leq$8km/h\\ (horizon: 10 to 50 minutes)
}
  &\shortstack[l]{2 ring roads\\Beijing (China)
} 
  &\shortstack[l]{insights into\\congestion source \\detection
} \\
  \hline

    \shortstack[l]{\citep{ma2015large} } 
  &\shortstack[l]{Traffic speed}
  &\shortstack[l]{RBM, RNN} 
  &\shortstack[l]{Accuracy 88.2\%}
  &\shortstack[l]{GPS data\\515 road links\\Ningbo, China} 
  &\shortstack[l]{Extensive sensitivity\\ analysis w.r.t \\ binary threshold \\on speed} \\
  \hline  

    \shortstack[l]{\citep{yu2017deep}} &
    \shortstack[l]{Traffic speed} &
    \shortstack[l]{LSTM} & 
    \shortstack[l]{MAPE: 5\% } &
    \shortstack[l]{2018 VLDs\\ (45 days)\\California, USA} &
    \shortstack[l]{ \\spatio-temporal \\analysis of \\performance} \\
    \hline
\shortstack[l]{\citep{sun2019traffic}} 
  &\shortstack[l]{ Traffic speed}
  &\shortstack[l]{CNN \\ LSTM} 
  &\shortstack[l]{90.55\%$\leq$Accuracy$\leq$96.32\% \\ 91.89\%$\leq$Accuracy$\leq$96.75\%  } 
  &\shortstack[l]{2000 taxis GPS\\(28 days)\\Chengdu, China} 
  &\shortstack[l]{extensive sensitivity \\analysis w.r.t \\input horizon} \\
  \hline

        \shortstack[l]{\citep{cheng2018deeptransport} } 
  &\shortstack[l]{Traffic speed 
}
  &\shortstack[l]{novel~\href{https://d3i71xaburhd42.cloudfront.net/56f17fa9a094e80f3e5e47311f2f31613d48e7b0/4-Figure4-1.png}{architecture}\\(built using \\CNN,LSTM \\ \& attention)}
 
  &\shortstack[l]{QWK 0.52 at 60 minutes }
  &\shortstack[l]{349 road links\\(4 months)\\Beijing (MapBJ)} 
  &\shortstack[l]{Insights into \\upstream and\\ downstream flows} \\
  \hline  
  
   \shortstack[l]{\citep{ranjan2020city} } 
  &\shortstack[l]{Traffic speed }
  &\shortstack[l]{novel~\href{https://www.researchgate.net/profile/Navin-Ranjan/publication/341062590/figure/fig4/AS:895246355398656@1590454772095/PredNet-model-architecture-a-The-model-is-divided-into-three-sections-section-A-is.png}{PredNet})\\(built using \\CNN\&LSTM)} 
  &\shortstack[l]{Accuracy: 84.2\%}

  &\shortstack[l]{Speed heat map\\ Seoul, \\S Korea
} 
  &\shortstack[l]{scalable\\ architecture
} \\
  \hline      

     \shortstack[l]{\citep{shin2020prediction} } 
  &\shortstack[l]{Traffic speed 
}
  &\shortstack[l]{LSTM
} 
  &\shortstack[l]{MAPE: 4.29\% (urban)\\MAPE: 6.08\%(suburban)
}
  &\shortstack[l]{Urban \\ suburban\\ areas in \\\& around Seoul,\\ S. korea
} 
  &\shortstack[l]{observation: variation in\\ complexity of task\\ based on the \\type of network
} \\
\hline

    \shortstack[l]{\citep{liu2017short} } 
  &\shortstack[l]{Traffic flow}
  &\shortstack[l]{Conv-LSTM
\\Conv bi-dir-LSTM
} 
  &\shortstack[l]{MAPE 9.53%
\\MAPE 6.98%
} 
  &\shortstack[l]{~\href{https://dot.ca.gov/programs/traffic-operations/mpr/pems-source}{PeMS}\\California, USA} 
  &\shortstack[l]{bi-directional\\ LSTM for\\ historical data
} \\
  \hline
  
  \shortstack[l]{\citep{li2017diffusion}} 
  &\shortstack[l]{Traffic flow}
  &\shortstack[l]{novel~\href{https://i.loli.net/2019/04/23/5cbec05ce557e.png}{DCRNN}\\(built using \\encoder \& decoder)}
  &\shortstack[l]{MAPE: 2.9\%,3.9\%,4.9\% \\ \hspace{0.9cm} for 15, 30, 60 \\minutes respectively }
  &\shortstack[l]{~\href{https://dot.ca.gov/programs/traffic-operations/mpr/pems-source}{PeMS}\\ California, USA } 
  &\shortstack[l]{incremental training\\using scheduled\\ sampling; insights \\into effects \\ of thresholds\\on load factor} \\
  \hline  

\shortstack[l]{\citep{rahman2020real}} & 
            
                \shortstack[l]{Queue length} &
            \shortstack[l]{LSTM} &         
            \shortstack[l]{RMSE{\raise.17ex\hbox{$\scriptstyle\sim$}}1 }&
                        \shortstack[l]{11  intersections (VLDs) \\3 months\\Florida, USA} &
                            \shortstack[l]{efficient encoding\\ for spatial \\information }\\
                                    


  \hline

\shortstack[l]{\citep{7966128} } 
  &\shortstack[l]{$\frac{average~speed}{speed~limit}$
}
  &\shortstack[l]{CNN, RNN
} 
  &\shortstack[l]{--
}
  &\shortstack[l]{~\href{pems.dot.ca.gov}{PeMS}\\ California, USA
} 
  &\shortstack[l]{Node-specific \\thresholds for\\ better generalisation
} \\
  \hline

               \shortstack[l]{\citep{di2019traffic} } 
  &\shortstack[l]{Not applicable \\(pre-labelled by \\data provider\\~\href{https://www.here.com/}{HERE} api)
}
  &\shortstack[l]{Conv-LSTM
} 
  &\shortstack[l]{MSE:0.73 (weekdays),\\~~~~~~0.37 (weekend)
}
  &\shortstack[l]{553 road links \\(5 weeks)\\ Helsinki, Finland
} 
  &\shortstack[l]{congestion \\tree
} \\
\bottomrule
    
    \end{tabular}
   \end{adjustwidth}
\end{table}

\textbf{Summary:} In this subsection, various aspects of deep learning applications in predicting recurring congestion are discussed. The takeaways from this subsection are summarised below:
\begin{itemize}

\item We observe that there have been very few attempts to comprehensively compare the performance of different deep learning models while keeping the dataset and the specific prediction task fixed. We know that the performance of a deep learning model is largely dependent on the choice of hyperparameters (such as the number of hidden layers, number of convolutions, learning rate etc.). When new deep learning architectures are presented for specific tasks, a fair comparison with previously used architectures might not be feasible. The reason behind this is that the best performance from a deep learning model involves training a large number of models to determine the best set of hyperparameters. So, it is not plausible to do such fine tuning for all the previous work.\\
The solution might be to establish public benchmarks for each dataset while keeping the test, train, validation split fixed. Such dataset specific benchmarks are widely popular in the computer vision community. They might significantly reduce the duplication of efforts to reproduce  results from a previous paper. Another benefit of such benchmarks would be to reduce the duplication of data pre-processing as it can be a more time consuming task than training a deep neural network iself. 

\item Particularly, while predicting congestion over a large part of the network for several time steps, we observe that extensive sensitivity analyses has been reported in some papers in order to reveal the temporal and spatial variation in performance. Highlighting such differences in performance has resulted in future work being targeted at improving the spatial and temporal generalisation of the deep learning models. We believe this is a good trend and future research should include more of such analyses. It also helps us to understand the limits of short-term traffic prediction using deep learning. Based on the papers presented here, the maximum time horizon of prediction appears to be 60 minutes.\\
We believe that future attempts to increase the prediction horizon will be useful to garner trust in the deep learning solutions for traffic congestion prediction.  

\item We observe a dearth of papers which present deep learning models that are updated as new data arrives. This is known as online learning. Online learning is a framework and not a model. So, theoretically any deep learning model can be integrated into an online learning framework and model updates can be demonstrated. So far, it is not popular because live traffic data are not easily available to researchers.\\
Even when streaming new traffic data are not available, online learning capabilities can still be demonstrated by using temporal splits of the historical data and evaluate model performance as it gets trained on increasing amounts of data.

\end{itemize}

\subsection{Deep learning for non-recurring congestion}\label{subsdlNrc}
An exhaustive list of causes behind non-recurring congestion is not known~\citep{mcgroarty2010recurring}. New causes are added as new data sources become available and new correlations and causalities are established. Some well investigated causes of non-recurring congestion are traffic accidents, varying weather conditions, disasters and planned events. In this survey, we focus on deep learning applications for predicting congestion due to accidents. The reasons for this focus on accidents are threefold. First, among this list of well-studied causes, traffic accidents are the leading cause behind a large percentage of non-recurring congestion~\citep{hallenbeck2003measurement}. Second, deep learning has been widely used to predict congestion after traffic accidents. Third, investigating traffic congestion due to other causative factors (weather, planned events and disasters) are best suited for scenario based studies using traffic simulators. Such studies are usually conducted in the planning stage and hence the high computation time of using traffic simulators do not present a challenge~\citep{aljamal2018comparison}. 
A literature search reveals that there is a wide variation in the specific deep learning task when using deep learning methods to predict congestion after traffic accidents. We have grouped the research into the following two clusters:
\begin{itemize}
    \item prediction of post accident traffic congestion
    \item prediction of post accident congestion clearance time 
\end{itemize}
The papers discussed in this subsection are summarised in~\autoref{tab:predict_nonrecurrent_congestion}.
\textbf{Post accident traffic congestion prediction}:~\cite{sun2017dxnat} propose a CNN based~\href{https://ieeexplore.ieee.org/mediastore_new/IEEE/content/media/8241556/8257893/8258162/8258162-fig-3-source-small.gif}{architecture} to predict traffic flow after accidents. The traffic speed data obtained from the traffic information system (\href{https://www.here.com/}{HERE api}) is  converted to traffic heat map images. If the speed ($v$) is less than 80 miles/h, the pixel value $p$ is set to ${\left(1-v/80\right)*255}$ and $0$ otherwise. A single threshold is used to define a binary classification task with two classes -- recurring and non-recurring congestion. The reported accuracy is 86.6\% with a false positive rate of 13.71\% and a false negative rate of 4.44\% (FN: model wrongly classifies non-recurring congestion as recurring congestion). An interesting contribution from their work is the use of crossover operator for reducing data imbalance issues (low number of `accident' data points compared to `no accident' data points). Crossover is a technique commonly used in genetics to model the creation of new chromosomes by partial exchanges of the genetic material of parent chromosomes. They hypothesise that various traffic data points collected within a short time range have the same event label and hence, applying  crossover to random locations in the traffic heat maps results in data augmentation without compromising the data quality.\\
~\cite{yu2017deep} propose a mixture model which has two components, one composed of LSTM and the other composed of an autoencoder. The incident data are fed to the autoencoder and the traffic data are fed to the LSTM. Finally, the outputs from the two components are concatenated and a fully connected layer is used to output traffic speed at each sensor location. They use vehicle loop detector (VLD) speed recordings from the publicly available \href{https://dot.ca.gov/programs/traffic-operations/mpr/pems-source}{Caltrans} dataset from California, USA. While predicting post incident traffic speed for a prediction horizon of 3 hours, their model achieves a MAPE of 0.97\%. For comparison, an LSTM model achieves a MAPE of 1.00\% and a three layered feed forward neural network achieves 3.65\%. An interesting contribution of their work is the use of signal stimulation to investigate the effects of abrupt reductions of  input speed on the model response. They report that the model's response remains unchanged when the stimulations last only for short durations (<5 minutes), thereby suggesting that the model is robust to the minor fluctuations in the input data. Additionally, they report that the model response is amplified when the stimulations are injected during peak hours.\\
~\cite{fukuda2020short} propose \href{https://ietresearch.onlinelibrary.wiley.com/cms/asset/d4bbf261-631b-47f8-92c0-de91bfc01499/itr2bf00897-fig-0004-m.png}{X-DCRNN}, which is an extension of DCRNN in order to input the incident data explicitly. Their dataset is created using simulations on the microscopic traffic simulator MATES~\citep{yoshimura2006mates}. The traffic demand was calibrated using the meta data obtained from the local transportation authority. The network in the simulator is based on the central business district of Okayama city in Japan, consisting of 206 traffic sensors across 339 road segments and spread across 3 square kilometres. Their model predicts post incidence traffic speeds and the errors are reported for incident segment and the corresponding downstream segments. For the  incident road segment, the reported MAE is 0.74 miles/h and the RMSE is 0.87 miles/h. For comparison, on the same dataset, the DCRNN model achieved a MAE of 1.97 miles/h and an RMSE of 5.64 miles/h. While predicting the traffic on the immediate downstream road segment, both DCRNN and  X-DCRNN achieve a similar level of performance. For immediate downstream segment, X-DCRNN achieves MAE of 3.68 miles/h and RMSE of 6.39  miles/h. For comparison, the DCRNN model achieved MAE of 3.81 miles/h and RMSE of 6.33 miles/h.\\

\textbf{Predicting congestion clearance time}: Congestion clearance time is a useful indicator of assessing the impact of accidents on traffic congestion. After an accident, traffic flow reduces due to restricted movement on affected lanes. Typically, the traffic flow eventually decreases to a minimum value and then recovers. When modelling congestion clearance time, most papers presented here focus on the time duration between the instant when the maximum level of congestion is reached (lowest flow) to the time instant when the flow returns back to pre-accident levels. A FFNN with two hidden layers is used in~\cite{do-part-me-kiya-hai-total-clearance}. They use average speed obtained from 173 vehicle loop detectors (VLDs) in Shanghai, China and accident data from traffic police records to create input vectors having 9 features. While predicting congestion clearance time, their model achieves a MAPE of 40\% and an RMSE of 8.3 minutes. For comparison, a multilinear regression model achieved a MAPE of 49.8\% and an RMSE of 10.22 minutes. The total number of accident records in the dataset was 4017. \\
~\cite{lin2020real} leverage the idea that in the instances of post accident congestion, the exact details about the accident are not available instantly. More information about the type, location, severity and affected lanes are available as the damage is assessed by the bystanders, involved parties or emergency response. They propose a framework capable of updating the prediction with the arrival of new information. They define non-recurring congestion prediction as a multiclass classification task (5 classes). The class of congestion is defined using thresholds on the maximum value of congestion delay index ($CDI~=~\frac{v_{free}}{v}$) reached after the accident. The congestion clearance time is defined as the time elapsed while the $CDI$ returns from its maximum value to pre-accident levels. They then predict the congestion clearance time for each type of accident. They propose a~\href{https://ars.els-cdn.com/content/image/1-s2.0-S0001457520305807-gr3.jpg}{framework}, consisting of FFNN with one hidden layer that is capable of updating the model using the new data available during the course of congestion clearance time. The model update feature results in a significant performance improvement for the most severe accident class: RMSE (minutes) decreases from 10.8 to 7.62 and MAPE decreases from 17.4\% to 9.33\%. Their data was collected using an anonymous navigation system from Beijing, China.\\
~\cite{li2020deep} propose an architecture called \href{https://ars.els-cdn.com/content/image/1-s2.0-S0952197620301226-gr1.jpg}{\textit{fusion} RBM} which is created by concatenating the outputs of two sets of stacked RBMs. The {fusion} aspect of their model is inspired by the fact that the accident data consist of categorical variables while the traffic data are continuous. One stacked RBM unit takes the categorical accident data as input while the other stacked RBM unit takes the continuous data as input. Finally, the outputs from both the units are concatenated and passed through a single neuron to finally output the congestion clearance time. The predicted congestion clearance time is quantified into ten levels using 10 minute increments (0-10,10-20 and so on). The reported MAPE is 20.23\% and the RMSE is 11.84 minutes. They used traffic data collected from California~(\href{https://dot.ca.gov/programs/traffic-operations/mpr/pems-source}{PeMS}) and traffic incidents data collected from a highway safety information system (HSIS). The total number of accidents in their dataset was 968 with a mean congestion clearing time of 37 minutes. \\ 

\textbf{Summary:} We discussed the applications of deep learning in predicting post accident traffic congestion and its clearance time. The key difference from recurring congestion prediction was that data from multiple sources must be {fused} in order to predict post accident congestion. Each paper presented here handled the challenge of data fusion in different ways, with almost no consensus between them. Benchmarking of standard data fusion algorithms from multiple sources might be helpful to provide insights to future researchers about the most efficient techniques for traffic data fusion. Data fusion has been extensively studied in the Internet of things (IOT) community. IOT research is focused on pervasive communication between different devices, hence efficient data fusion has been researched extensively. Transportation researchers can draw inspiration from such sources and explore the possibilities of improving post accident congestion prediction using efficient data fusion techniques. 
~\cite{SURVEY-MULTIMODEL-DATA} compare the data fusion performance of different deep learning architectures. Most of the architectures presented in their survey are commonly used for deep learning based congestion prediction task, hence it would be interesting to use the insights presented therein. Predicting post accident traffic congestion can also be understood in the light of the traffic hysteresis curve presented in~\autoref{relev3}.



 \begin{table}[]
\begin{adjustwidth}{-1cm}{}
        \caption{\normalfont{Table summarising the papers discussed in~\autoref{subsdlNrc}} ({Deep learning for non-recurring congestion prediction})}
    \label{tab:predict_nonrecurrent_congestion}
    
    \begin{tabular}{l l l l l}
      \toprule
   Paper 
  & \shortstack[l]{DNN \\ architecture }  
  &  \shortstack[l]{ Performance} & \shortstack[l]{Data\\source} 
  & \shortstack[l]{Unique\\aspect}  \\
  \midrule

 \shortstack[l]{\citep{sun2017dxnat}}
  &\shortstack[l]{CNN} 
  &\shortstack[l]{Accuracy = 86.6\% \\FPR = 13.71\%\\ FNR = 4.44\% }
  &\shortstack[l]{Traffic speed data\\ from \href{https://www.here.com/}{HERE api}\\ 
  (6 days: train, \\1 day:validation)}
  &\shortstack[l]{Data augmentation\\using crossover }\\

  \hline
 \shortstack[l]{\citep{yu2017deep}}
  &\shortstack[l]{Mixture model\\ (LSTM and\\ autoencoder)}
  &\shortstack[l]{MAPE = 0.97\% \\(predicting post\\ accident traffic speed)}
  &\shortstack[l]{2018 VLDs\\ (45 days)\\California, USA}
  &\shortstack[l]{Robust model \\tested with \\stimulation response}\\

  \hline
 \shortstack[l]{\citep{fukuda2020short}}
  &\shortstack[l]{X-DCRNN\\extension of \\DCRNN~\citep{DCRNNWALAPAPER}}
  &\shortstack[l]{MAE = 3.68 miles/h\\RMSE = 6.39 miles/h\\(predicting post\\ accident traffic speed)} 
  &\shortstack[l]{simulated data\\for 206 traffic sensors,\\ 339 road segments \\generated using MATES\\ (calibrated demand \\from Okayama, Japan) }
  &\shortstack[l]{{Large number of} \\ training data points\\ possible due to \\simulated data} \\
  
  \hline
 \shortstack[l]{\citep{do-part-me-kiya-hai-total-clearance}}
  &\shortstack[l]{FFNN \\(2 hidden layers)}
  &\shortstack[l]{MAPE = 40\% \\
RMSE = 8.3 minutes} 
  &\shortstack[l]{173 VLDs from Shanghai\\4017  accident records \\from police records}
  &\shortstack[l]{Performance comparable\\ to multilinear regression\\ suggesting the use\\ of other ANN architectures } \\
  
  \hline
 \shortstack[l]{\citep{lin2020real}}
  &\shortstack[l]{FFNN \\(one hidden layer)}
  &\shortstack[l]{ MAPE = 9.33\%\\RMSE = 7.62 minutes}
  &\shortstack[l]{Anonymous Navigation\\ System\\Beijing, China}
  &\shortstack[l]{Novel data fusion\\ and model update\\ with new data} \\
  
  \hline
  \shortstack[l]{\citep{li2020deep}}
  &\shortstack[l]{specialised \\
\href{https://ars.els-cdn.com/content/image/1-s2.0-S0952197620301226-gr1.jpg}{fusion RBM}}
  &\shortstack[l]{MAPE = 20.23\% \\RMSE = 11.84 minutes.}
  &\shortstack[l]{968 accidents from~\href{https://www.hsisinfo.org/}{HSIS}\& \\
  traffic data from\\\href{https://dot.ca.gov/programs/traffic-operations/mpr/pems-source}{PeMS} dataset\\ California, USA}
  &\shortstack[l]{specialised architecture\\for data fusion}\\
  
\bottomrule
    
    \end{tabular}
\end{adjustwidth}
\end{table}

\section{Deep learning for congestion alleviation}
The congestion alleviation techniques differ significantly for recurring and non-recurring congestion. On the one hand, recurring congestion is caused due to infrastructural bottlenecks which are insufficient to handle the peak demand of traffic. So, the deep learning solutions for recurring congestion are targeted  at decreasing the severity of the recurring congestion by distributing the demand in an  optimal fashion. On the other hand, non-recurring congestion is caused primarily due to accidents. So, the deep learning applications for alleviating non-recurring congestion are targeted at reducing accidents. Deep learning has been widely used to predict the accident risk. The predicted accident risk can be used to alert drivers or to impose speed restrictions in order to reduce accidents. At the end of this section, we discuss the potential connection between the efforts to reduce recurring congestion and the efforts to reduce non-recurring congestion. As discussed in the previous~\autoref{subsdlrc}, other causes of non-recurring congestion such as planned events, bad weather and natural disasters are best suited for scenario based studies using traffic simulators and hence, deep learning has not extensively applied for non-recurring congestion due to those causes.

\subsection{Deep learning for recurring congestion alleviation}\label{recurringalleviate}
As soon as recurring congestion is predicted by a congestion prediction model, alleviation measures can be put in place to reduce the build up of congestion. At the network level, this can be achieved by controlling the supply parameters such as traffic signal control at intersections, ramp metering, imposing speed limit restrictions and imposing lane use restrictions. At the individual level, this can be achieved by using descriptive, prescriptive or hybrid  methods. Descriptive methods involve broadcasting vital information about the network traffic conditions to the drivers and helping them make an informed decision about their trip start time and chosen route. Prescriptive methods involve trip specific suggestions to drivers (such as optimal start time and optimal route choice). Hybrid methods use a combination of descriptive and prescriptive methods. The descriptive and prescriptive methods face significant challenges due to the additional layer of human behavior modelling in those approaches. In this survey, we focus on the recent research using deep learning for congestion alleviation at the network level. \\

\textbf{Challenges in using demand side solutions:} The challenges encountered when using descriptive and prescriptive methods for congestion alleviation are presented in~\cite{moshekapaper}. The biggest challenge is the information flow back from the drivers to the network. The information is crucial in order to estimate the effect of the suggested route on the roads. They strongly argue in favour of the need for large-scale tracking data. Due to the high penetration of smart phones in urban areas, such real time traffic information is increasingly being relayed to the drivers through their mobile phones. The challenge however arises from the fact that the descriptive and prescriptive solutions are being grouped together. When a user queries for driving directions, the preferred routes returned by the navigation application incorporates some route suggestions. The final goal of such personalised route suggestions is not known, since the companies which provide such services present mobility as a service and the  algorithmic details are trade secrets. A survey conducted in 2018 in the US reports that 87\% of drivers use some navigational application to get suggestions for driving  directions~\citep{travel-survey-87-percent}. A report released by Google Maps states that more than a billion kilometres of travel per day are tracked using their application~\citep{google-ka-report}. When the coverage and compliance of such personalised route suggestions increase, it can potentially alter the user equilibrium in the transportation system. Since such efforts from the private sector are targeted at creating route choice as a service and the exact methods are black boxed as trade secrets, it is difficult to estimate the potential drawbacks related to the equity of such solutions. In the absence of real data, researchers have used economic instruments as an attempt to increase the compliance and hence, ease the task of modelling the human behavior. The compliance of advisory measures can be increased using punitive measures or rewards. The choice between which of the two approaches proves to be more effective, is a difficult one to make. The effectiveness of punitive measures is similar across various geographical locations whereas the effectiveness of reward based measures has been found to vary significantly~\citep{tillema2013charging,li2019impacts}. Two commonly used punitive measures are congestion tolling and tradable tokens.~\cite{de2018congestion} present a methodology for comparing the effects of both approaches and strongly argue in favour of tradable tokens. Various ways in which new and emerging technology can be used to achieve efficient congestion tolling are discussed in~\cite{de2011traffic}. One of their suggestions is automatic number plate recognition (ANPR). Deep learning has been widely applied for ANPR. However, this application is not covered in this survey. The interested reader is referred to~\cite{connie2018review} and~\cite{khan2019survey}. The challenges encountered in applying ANPR to developing countries are presented in~\cite{kyaw2018license}. \\


\textbf{Supply side solutions at the network level:}
 We focus on the deep learning applications for congestion alleviation at the network level. The majority of such efforts use deep reinforcement learning. As discussed in~\autoref{section:machineTrans},  a reinforcement learning framework has four main components: agent, action, reward and the environment. The concept of \textit{policy} fits well into the traffic signal control problem. The best policy (\textit{e.g.}, order of the red and green lights in different directions at an intersection) can be found by optimising the model for which the reward (\textit{e.g.} maximum cumulative flow in the network) is minimised. In the recent research, deep Q-network (DQN) is being commonly used and authors are increasingly interested in investigating the equity aspect of such solutions. The traffic controlling infrastructure (such as traffic signal or speed limit signs) is modelled as the agent. The choice of agents determines which actions are allowed. The papers are arranged into two headings based on the chosen agent.\\

\begin{itemize}
\item Adaptive Traffic Signal Control (TSC)
\item Variable  Speed  Limit  Control  (VSLC)
\end{itemize}
The papers discussed in this subsection are summarised in~\autoref{tab:alleviate_recurrent_congestion}

 \begin{table}[]
\begin{adjustwidth}{-2.2cm}{}
        \caption{\normalfont{Summary of papers discussed in~\autoref{recurringalleviate}} ({Supply side solutions using deep learning for recurring congestion alleviation})}
    \label{tab:alleviate_recurrent_congestion}
    
    \begin{tabular}{l l l l l l}
      \toprule
   Paper 
  & \shortstack[l]{DNN \\ architecture }  
  &\shortstack[l]{Reward \\function}
  &  \shortstack[l]{Performance} &\shortstack[l]{Data\\source} 
  & \shortstack[l]{Unique\\aspect}  \\
  \midrule

 \shortstack[l]{\citep{genders2016using}}
  &\shortstack[l]{CNN based DQN} 
  &\shortstack[l]{cumulative delay\\queue length}
  &\shortstack[l]{cumulative delay\\ = 719 seconds\\queue length = 13}
  &\shortstack[l]{small network\\(one intersection \\with 4 segments\\each with 4 lanes)}
  &\shortstack[l]{identified several \\important challenges \\ (such as fairness)}\\

  \hline
 \shortstack[l]{\citep{yen2020deep}}
 &\shortstack[l]{CNN based DSARSA}
  &\shortstack[l]{Power metric =\\ $\frac{maximum~throughput}{end-to-end~delay}$}
  &\shortstack[l]{Power metric = 35}
  &\shortstack[l]{9 intersections\\ (3X3 grid)}
  &\shortstack[l]{Novel reward function\\DQN convergence \\challenges highlighted}\\

  \hline
 \shortstack[l]{\citep{zeng2018adaptive}}
  &\shortstack[l]{LSTM based DQN}
  &\shortstack[l]{Average waiting\\ time} 
  &\shortstack[l]{Average waiting\\ time = 17.71 s}
  &\shortstack[l]{small network\\ (one intersection \\with 4 segments) }
  &\shortstack[l]{Performance stable\\ at low\\ penetration ratios\\<50\%}\\
  
  \hline
 \shortstack[l]{\citep{genders2019asynchronous}}
  &\shortstack[l]{FFNN based DQN}
  &\shortstack[l]{end-to-end-delay} 
  &\shortstack[l]{--} 
  &\shortstack[l]{small network \\(one intersection\\ with 4 segments)}
  &\shortstack[l]{40\% improvement\\ over actuated \\signal control} \\
  
  \hline
 \shortstack[l]{\cite{7795890}}
  &\shortstack[l]{(classical RL)}
  &\shortstack[l]{--}
  &\shortstack[l]{--}
  &\shortstack[l]{--}
  &\shortstack[l]{Priority vehicles\\considered} \\
  
  \hline
  \shortstack[l]{\citep{shabestray2019multimodal}}
  &\shortstack[l]{CNN based RL}
  &\shortstack[l]{cumulative delay \\at intersections}
  &\shortstack[l]{For transit only\\
  average TT :292 s\\compared to 343 s \\for actuated control}
  &\shortstack[l]{6 intersections\\Ontario, Canada\\(calibrated demand)}
  &\shortstack[l]{1. Multi-modal\\ signal controller\\ 2. Calibrated demand\\3. Head-count \\based priority}\\
  
    \hline
  \shortstack[l]{\citep{wu2020a}}
  &\shortstack[l]{LSTM based DQN}
  &\shortstack[l]{several components\\including queue \\lengths, mean vehicle\\delay, number of\\pedestrians delayed}
  &\shortstack[l]{During peak hours\\50\% reduction in\\ queue length\\ compared to\\ other baselines}
  &\shortstack[l]{multiple intersections\\simulator: SUMO}
  &\shortstack[l]{Pedestrian waiting\\ incorporated into\\reward function}\\
  
      \hline
  \shortstack[l]{\citep{croatia-CNN-DQN}}
  &\shortstack[l]{CNN based DQN}
  &\shortstack[l]{three components\\ 1. traffic flow \\2. safety constraints\\3. driver comfort}
  &\shortstack[l]{13.5\% increase \\ in average speed \\25\% decrease\\ in density\\compared to\\no vehicle \\speed limit (VSL case)}
  &\shortstack[l]{8 km of highway\\simluator: VISSIM}
  &\shortstack[l]{Novel reward function }\\

  \hline
  \shortstack[l]{{\citep{wu2020b}}}
  &\shortstack[l]{{FFNN based DRL}}
  &\shortstack[l]{four components\\1. increasing flow\\ 2. decreasing travel\\ time\\3.reducing rapid\\ acceleration\\4. reducing emissions}
  &\shortstack[l]{5.8\% decrease \\ in average travel time\\compared to\\no vehicle\\ speed limit (VSL case)}
  &\shortstack[l]{850 m of highway\\simulator: SUMO}
  &\shortstack[l]{Assuming autonomous \\ vehicles and \\allows for \\abrupt changes\\in speed limit\\and lane \\specific limits}\\

\bottomrule
    
    \end{tabular}
\end{adjustwidth}
\end{table}

\subsubsection{Adaptive traffic signal control (TSC)}\label{TSCsection}
A CNN based DQN was presented in~\cite{genders2016using} in order to model the signal controlling agents in a traffic simulator. The goal was to maximise the \textit{network-wide} throughput by optimising the behavior of these agents. They analyse the performance of their model based on three parameters -- the cumulative delay, the queue length at intersections, and the average travel  time. They model one intersection with four segments from four directions each having four lanes. The vehicles are loaded at varying flow rates between  0 to 450 vehicles/h. The left and right turning traffic is loaded using the inverse Weibull distribution~\citep{Weibull} while the through traffic is loaded using the Burr distribution~\citep{burr}. Using CNN based DQN to control the actions of the traffic controlling agent, they achieve a cumulative delay of 719 seconds and an average queue length of 13 vehicles. For comparison, a DQN based on FFNN with one hidden layer achieves an average queue length of 33. They highlight the need to study the \textit{fairness} aspect of the policy optimised by the DQN. In the absence of a {fairness} metric of the model, it is possible that the algorithm favors or disfavors specific movements of traffic. They assert that the fairest policy might not result in the best throughput of traffic -- hence, there is a need to search for a balance between the two objectives by tweaking the reward function.~\cite{yen2020deep} attempt to address this difference of objectives by incorporating a \textit{power metric} in their reward function. The {power metric} is defined as the ratio of maximum throughput to the end-to-end delay ($\frac{flow}{time}$). When the power metric is maximised, the throughput (the flow of traffic) is maximised and the end-to-end delay (travel time) is minimised. Their network consists of nine intersections (3x3 grid). Using power metric as the reward function however did not result in  convergence of the DQN models. Hence, they used DSARSA instead of DQN to achieve convergence. SARSA stands for State-Action-Reward-State-Action. It was introduced in~\cite{sarsaPaper} and differs from DQN in the choice of the next action. The SARSA algorithm uses the same policy to choose the next action as well as update the reward function whereas the DQN uses a different (greedy) policy to choose the next action. Using their CNN based DSARSA, they achieved a power metric of 35. \\ Another attempt in this direction is~\cite{zeng2018adaptive} who propose DRQN (Deep Recurrent Q Learning) using LSTM based DQN. They report an average waiting time of 17.71 seconds when using DRQN on their synthetic network consisting of a single four-way intersection. The distribution used for loading the vehicles is binomial. The performance of LSTM based DQN is similar to that of a CNN based DQN as used in \cite{genders2016using}. However, the performance of the LSTM-DQN model is stable when the penetration ratio is varied. Penetration ratio is defined as the percentage of vehicles on the road who shared their data. The CNN-based models perform poorly when the penetration ratio falls below $0.5$. At a penetration ratio of $1$ (implying data from all vehicles are available), the performance is similar for both  models.~\cite{genders2019asynchronous} use a feed forward neural network (FFNN) based DQN to optimise the traffic control policy during peak hours. Using their synthetic network comprising a single intersection, they report a reduction of around 40\% in the average total vehicle delay when compared to the traditionally used actuated method of traffic signal control~\citep{newell1989theory}.\\

Most of the research discussed in this paragraph uses a traffic simulator to observe the effect of change in traffic-controlling policy at each intersection. We observed that SUMO~\citep{behrisch2011sumo} is the most commonly used traffic simulator for this purpose. The main reason behind the popularity of SUMO is that it has a python interface. Python is the most commonly used language for deep learning research, hence the integration with SUMO is smooth. \\

 Fairness is an important factor for  deep learning based solutions for TSCs in order to ensure wide acceptance of such solutions by the transport-planners and policymakers. When priority vehicles are taken into account, the importance of {fairness} increases significantly. Priority vehicles include transit vehicles like buses and emergency transportation like ambulances. Researchers have tried to incorporate fairness constraints into the reward function of reinforcement learning solutions.~\cite{7795890} used a classical (not using neural network) reinforcement learning algorithm and modified their reward function so as to penalise the signal controlling agent each time a priority vehicle is queuing at an intersection. Similar penalties have been adopted by the deep learning community in order to demonstrate fairness in the deep learning based reinforcement learning solutions. ~\cite{shabestray2019multimodal} propose a multi-modal signal controller using DQN and analyse the effects on various modes of transportation. Unlike other studies summarised in this subsection, this is the only paper which used real data to calibrate the traffic demand. They modelled a road network with 6 intersections representing a busy part of a city in  Ontario, Canada using the commercially available traffic simulator~\href{https://www.paramics.co.uk/en/}{PARAMICS}. Instead of prioritising vehicles specifically, they use a person-based performance of the controller in their reward function. This automatically prioritizes vehicles with higher head-count, implying that a bus has a higher priority than a car; a fully-occupied bus has a higher priority than a partially-occupied bus. The downside is that high-priority but low head-count vehicles like ambulances are at a disadvantage when this approach is used.~\cite{wu2020a} used a synthetic network with~\href{https://ieeexplore.ieee.org/mediastore_new/IEEE/content/media/25/9166810/9103316/wu2-2997896-small.gif}{multiple intersections} modelled in SUMO and proposed an LSTM based DQN to assign priority to vehicles and also include the pedestrian waiting time in their reward function. This concept of using power metric in the reward function is inspired by the literature on wireless networks \citep{kleinrock2018internet}. \\


\subsubsection{Variable  Speed  Limit  Control  (VSLC) } 
Similar to TSC, the VSLC methods are targeted at reducing the severity of an impending traffic congestion. The specific objective is to maximise the total flow in the network while minimising the average travel time. The design of the reinforcement learning framework for VLSC is similar to that of TSC. Here, the VSLC signs are modelled as the action taking agents. The actions allowed by the VSLC agent are to post variable speed limit signs on the variable message sign board (VMS). A microscopic traffic simulator is used to evaluate the effects of the action taken by the VSLC agent. However, TSC and VSLC differ significantly in their potential to be implemented in the real world. The challenge in the real-world implementation of VSLC stems from the potential lack of compliance on the part of drivers. Hence, in order to ensure higher compliance, driver comfort must be taken into account when formulating VSLC based solutions. In comparison, driver compliance is less of a concern in case of TSC based solutions because in practice, the instances of traffic light violation are far lower than traffic speed violations~\citep{wu2021mid}. The research summarised here has attempted to incorporate driver comfort metrics when using deep reinforcement learning models to find the optimal VSLC. \\

A three layered CNN based DQN model is presented in~\cite{croatia-CNN-DQN}. Their reward function is a weighted sum of three components. The first component is aimed at increasing traffic flow. The second component ensures safety by imposing constraints on the magnitude of variations in VMS posted speed limits. The third component ensures driving comfort by minimising the oscillations in the VMS posted speed limits. Their~\href{https://ieeexplore.ieee.org/mediastore_new/IEEE/content/media/9212529/9219008/9219031/gregu4-017_orig-research-large.gif}{network} was modelled using VISSIM microscopic traffic simulator and consists of eight kilometres of highway with one variable message sign board (VMS). The synthetic demand was sufficient to create 20 minutes of simulated congestion on the network. Using their DQN model, they demonstrate a 13.5\% increase in the average speed and 25\% decrease in the average traffic density.~\cite{wu2020b} consider a futuristic scenario where the connected and autonomous vehicles are ubiquitous. In such a scenario, the high compliance factor allows the VSLC agent to make more dynamic changes. They present a differential vehicle speed limit model, by applying lane specific variable speed limits. They train four different models for different rewards: increasing flow, decreasing travel time, reducing rapid acceleration and reducing emissions. The traffic demand used in this study is more realistic compared to previous work. The~\href{https://ars.els-cdn.com/content/image/1-s2.0-S0968090X20305647-gr2.jpg}{network} is a representation of 850 metres of a major highway from California and the demand used was calibrated using the data available for the highway. The mode composition was 85\% passenger vehicles and 15\% buses and trucks. Compared to the no-VSL case, their DVSL model achieves a 5.8\% reduction in average travel time. They used SUMO as their microscopic traffic simulator and include specific details about how different  functionalities in the SUMO application interface can be utilized to implement various components of the reward function.\\

\textbf{Summary:} We discussed the applications of deep learning when applied to limit supply using network level control of traffic. We observe two major challenges in the current state of research in this field as summarised below:
\begin{itemize}
    
\item The small size of road network and the homogeneity of traffic stream used for these studies poses a significant challenge to the validation of the research as a deployment option. The traffic demand is created using mathematical distributions rather than using the real traffic demand. Only one of the papers presented here used real data to calibrate their demand. Secondly, the reported performance of the DQN models is often not compared with the existing methods of network control (such as fixed phase transitions, Actuated). Instead, the comparison between DQNs using different neural network architectures is common. Unless the DQN based solutions are demonstrated to perform better than the existing methods, the real-world deployment of such solutions might be difficult. 
\item Another aspect discussed here was the lack of deep learning based solutions for managing the travel demand. The majority of large-scale solutions are provided by the private sector and the academic research is limited by data availability. Personal mobility data when fused with other sources, can reveal a lot of information about the travellers, hence the hesitation of the travellers public in sharing data is justified. We believe that a wide adoption of secure data pipelines in the research community might be an important step in garnering trust among the users who share data. Privacy preserving data fusion techniques have been explored in the Internet of things (IOT) community. A summary is presented in~\cite{privacypresen}. Transportation researchers can draw inspiration from such sources and adopt the best practices for privacy preservation. 
\end{itemize}


\subsection{Deep learning for non-recurring congestion alleviation}\label{non-recurringalleviate}
Accidents are the leading cause of non-recurring congestion. Hence an accurate prediction of the short-term accident risk enables us to proactively alert drivers about the same and potentially instruct them to reduce speed~\citep{8569437}. Recently, deep learning has been widely used to predict accident risk. In order to proceed with accident risk prediction, the common practice is to divide the study area into grids and then predict the number of accidents in each grid at different time intervals. The accident risk prediction can be designed as a binary prediction task where the deep learning model is trained to predict a label (`high risk' or `low risk') for each grid. It can also be designed as a regression task where the deep learning model is trained to predict the potential number of accidents in the grid. The papers discussed in this subsection are summarised in~\autoref{tab:alleviate_nonrecurrent_congestion}. \\

 \textbf{Demonstrating transfer learning capabilities}~\cite{najjar2017combining} use satellite imagery as inputs to their model. They use a model similar to AlexNet~(\autoref{cnnwalafigure}) in order to predict the accident risk into three levels.~\cite{8803732} and~\cite{najjar2017combining} have demonstrated transfer learning capabilities for the task of accident risk prediction. They trained their deep learning models using the data from one city and used the trained model to predict the accident risk in a different city (New York $\rightarrow$ Denver; London $\rightarrow$ New York).~\cite{8803732} report accuracies of 74.77\% and 76.20\% when using the model, which was trained on London data, to predict on data from New York and Denver respectively. Similarly, \cite{najjar2017combining} trained their model on data from New York and obtained an accuracy of 78.2\%. They then report comparable performance (73.1\%) when using the model from New York to make predictions on data from Denver.\\
 
 
\textbf{Spatial correlations and heterogeneity of the road networks:} The road network characteristics, the drivers' behavior and the travel demand vary spatially. On the one hand, this spatial heterogeneity results in a spatial variation in the prediction performance of deep learning models. On the other hand, it presents opportunities to leverage on the high correlation between data collected from sensors which were located in each other's vicinity.\\
In~\cite{zhao2018driving}, the authors apply principal component analysis (PCA) on the traffic and accidents data and then use a feed forward neural network with two hidden layers to predict the risk of a crash. Their data are based on one year of car accident records in the UK. PCA is used to identify un-correlated components in a dataset. The authors advocate the use of PCA due to the fact that in a transportation data set, many associated sensors and vehicles produce data which are correlated to some extent. Their experimental results  show a small increase (\char`\~1\%) of accuracy in crash risk prediction when using PCA compared to when not using it. \\
~\cite{garcia2018predicting} used a FFNN with one hidden layer to predict the accident risk. They defined three levels of accident severity and predict the risk for each of the three levels. Their dataset is based on 2 years of traffic and incident data from 13109 road segments in Switzerland. The data from each accident were encoded into 10 features and used as input. Their sensitivity analysis reveals that if different FFNNs are trained for different road types (\textit{e.g.} highways and tunnels), the average performance is better than using a single FFNN for the entire network with road type as one of the inputs.\\
In~\cite{yuan2018hetero}, the authors present Hetero-ConvLSTM, which, as the name suggests, is an extension of the ConvLSTM~ \citep{xingjian2015convolutional} architecture. The authors acknowledge the heterogeneous characteristics of the road network and hence, leverage it to produce better performances compared to previous work. They train different neural networks for different parts of the city and then create an ensemble to predict city-wide accident risk. They report an MSE of 0.021 for urban roads and 0.006 for rural roads. For comparison, the LSTM model achieves 0.187 and 0.042 for rural and urban roads respectively. Their dataset consists of 8192 grids with each having an area of 5x5=25 square kilometres. It comprises 8 years of data from various sources (weather, traffic, accidents, satellite) in Iowa, USA. The issue of spatial heterogeneity has been taken into account by other research. Some examples are~\cite{huang2019deep}, who propose a hierarchical fusion network to handle the temporal, and spatial heterogeneity of data and~\cite{zhou2019stack}, who use an attention  model to preferentially weight the network similarity factors to predict hourly crash risk.~\cite{8803732} use an attention model to handle the spatial heterogeneity and report an accuracy of 86.21\% using traffic data from London. \\

 \textbf{Online learning: }~\cite{fan2019research} propose an \textit{online} FC Deep learning model to predict accident blackspots. They report accuracies for three different types of weather, i.e. 88.70\%, 85.19\% and 83.33\% for sunny, cloudy and windy days respectively.

 \small{
 \begin{table}[h]
\begin{adjustwidth}{-2.3cm}{}
        \caption{\normalfont{Summary of papers discussed in~\autoref{non-recurringalleviate}} ({Deep learning for non-recurring congestion alleviation})}
    \label{tab:alleviate_nonrecurrent_congestion}
    
    \begin{tabular}{l l l l l l}
      \toprule
   Paper 
  & \shortstack[l]{DNN \\ architecture }  
  &\shortstack[l]{Predicted \\variable}
  &  \shortstack[l]{Performance} 
  &\shortstack[l]{Data\\source} 
  & \shortstack[l]{Unique\\aspect}  \\
  \midrule

 \shortstack[l]{\citep{najjar2017combining}}
  &\shortstack[l]{AlexNet, 3 levels} 
  &\shortstack[l]{Multi-class classification}
  &\shortstack[l]{73.1\%}
  &\shortstack[l]{Satellite imagery}
  &\shortstack[l]{Transfer learning\\trained on New York\\validated on Denver}\\
\hline

 \shortstack[l]{\citep{8803732}}
  &\shortstack[l]{ResNet} 
  &\shortstack[l]{Multi-class classification}
  &\shortstack[l]{74.77\% for NY\\ 76.2\% for Denver}
  &\shortstack[l]{Satellite imagery}
  &\shortstack[l]{Transfer learning\\trained on London\\validated on \\NY and Denver}\\
\hline
  
 \shortstack[l]{\cite{zhao2018driving}}
  &\shortstack[l]{FFNN \\(2 hidden layers)} 
  &\shortstack[l]{Crash risk \\prediction}
  &\shortstack[l]{--}
  &\shortstack[l]{1 year car accidents\\data from UK}
  &\shortstack[l]{Used PCA to\\ handle correlated\\readings from \\nearby sensors}\\
\hline

 \shortstack[l]{\citep{garcia2018predicting}}
  &\shortstack[l]{FFNN\\ (one hidden layer)} 
  &\shortstack[l]{Accident risk for \\different severity\\levels }
  &\shortstack[l]{MAPE \\(based on accident type)\\
  Light: 22·40\% \\
  Severe: 27·00\% \\
  Fatal: 30·00\% }
  &\shortstack[l]{2 years\\13109 road segments\\from Switzerland}
  &\shortstack[l]{Demonstrate that an \\ensemble of models\\each trained for \\ a single road type\\ outperforms\\ global model }\\
\hline

 \shortstack[l]{\citep{yuan2018hetero}}
  &\shortstack[l]{(Hetero) ConvLSTM} 
  &\shortstack[l]{Accident risk}
  &\shortstack[l]{MSE = 0.021 (urban)\\MSE = 0.006 (rural)}
  &\shortstack[l]{8192 grids \\(each 25 sq. km.)\\Iowa, USA}
  &\shortstack[l]{1. Leveraged heterogenity\\of road networks\\ to propose an ensemble\\2. Insights into \\differences between \\the complexity\\ of urban vs rural}\\
\hline

 \shortstack[l]{\citep{fan2019research}}
  &\shortstack[l]{FFNN} 
  &\shortstack[l]{Accident black spots}
  &\shortstack[l]{Accuracy \\(based on weather type)\\
  Sunny: 88.70\% \\ Cloudy: 85.19\% \\ Windy: 83.33\%}
  &\shortstack[l]{6391 accident records\\ over 18 months\\ from Jiangsu\\ Province China}
  &\shortstack[l]{1. Online learning\\ 2. Highlights importance\\ of weather\\ conditions}\\

\bottomrule
    
    \end{tabular}
\end{adjustwidth}
\end{table}
}




\subsection{Potential link between non-recurring and recurring congestion alleviation}
\begin{figure}[h]
\centering
\includegraphics[width=0.4\textwidth]{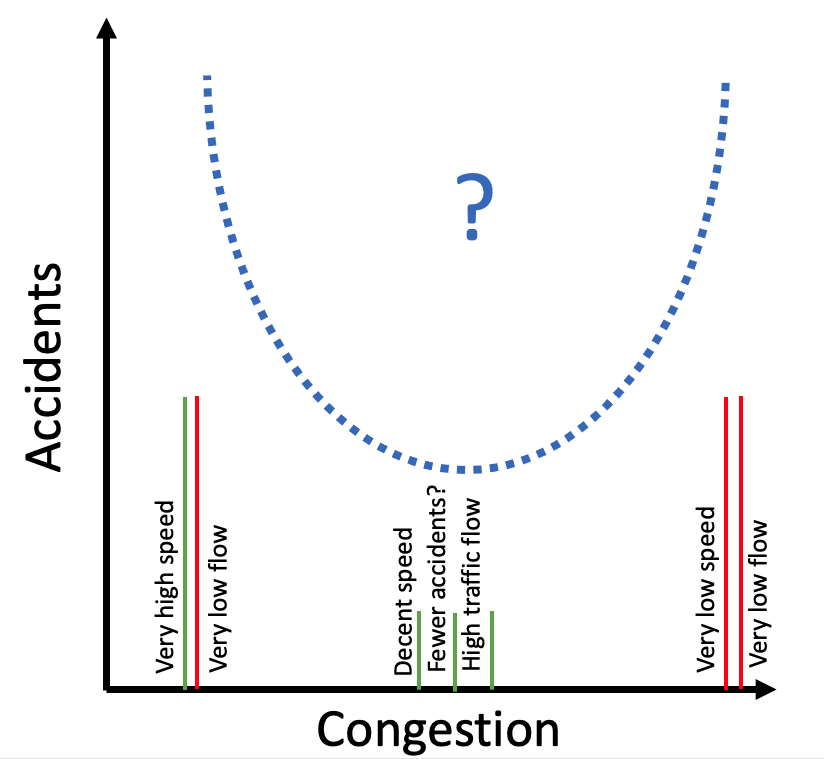}
\captionsetup{margin=4cm}
\caption{\texttt{A hypothetical U-shaped curve between accidents and congestion. Such curves have been observed in recent research.}}

\label{ushapedcurveqt}
\end{figure}

 The statement `\textit{accidents cause congestion}' is intuitive and undisputed but the reverse causality has not been established in a universal fashion.~\cite{retallack2019current} present a detailed summary of the current understanding of the relationship between traffic congestion and accidents. They report that researchers are far from reaching a consensus on the question, with conflicting reports from studies based on different types of roads and different countries of study. However, they find that a U-shaped curve between congestion levels and accidents appears to be a common observation in recent work which used a sufficiently large number of data points (\autoref{ushapedcurveqt}). A U-shaped curve implies that very high and very low levels of traffic density result in a higher number of accidents. If the U-shaped curve can be reproduced in future research, this can have far-reaching consequences in the attempts to reduce accidents and congestion. As shown in the~\autoref{ushapedcurveqt}, it would imply that the efforts to reduce accidents would also reduce congestion to levels of maximum flow of traffic~\citep{pasidis2019congestion}. In addition to the number of accidents, another important factor to consider is the severity of accidents. Establishing the causality behind the severity of accidents is a topic covered in road safety studies and is outside the scope of this survey. The interested reader is referred to \citep{wang2013effect}. Here, we have included papers that use deep learning methods for the prediction of accident severity. 

\section{Challenges and future directions}

\subsection{Challenges}
In this section, we summarise the challenges which were apparent while compiling the research papers for this survey. These challenges have been discussed in detail in the summary of specific sections. 
\subsubsection{Identifying the state-of-the-art model}
 In order to establish the state-of-the-art, the relevant work should be given a fair comparison. Fairness can be ensured by comparing the performance using just one parameter at a time -- while keeping all the other parameters unchanged. Ensuring such fairness is a challenging task in any field of research. Here, we discuss the challenges specific to the prediction tasks discussed in this survey.
\begin{enumerate}
    \item \textit{Different definitions of congestion based on different traffic variables:} As  discussed in~\autoref{defining-congestion}, congestion can be defined using different traffic variables. Due to the lack of a unified relationship between these variables, such differences in the choice of variables make a comparison between the reported prediction performances difficult. 
    
    

    \item \textit{Different quantisations of the same traffic variable:} Even if two research papers use the same traffic variable to define congestion, their quantisations might differ. For example, if the congestion is defined based on speed, one paper might present a binary prediction task (`{jam}' or `{no jam}'), while the other might present a multi-level variable (`{low traffic}', `{heavy traffic}' , `{total jam}'). If the dataset is fixed, a binary prediction task is \textit{usually} easier to solve than a multi-class prediction task \citep{allwein2000reducing}.

    \item\textit{Different resolutions of data across countries:} Researchers from different countries might work on the same prediction task but it may be unfair to compare their performance levels due to the differences in  data resolution. Such differences in resolution can be attributed to the infrastructure needed for data collection. Data collected from static sensors on roads in developed countries might have a higher resolution in space as well as time. The difference in the resolution of data implies a difference in the complexity of the prediction task at hand. Whether an increase in the resolution of data makes a prediction task easier or more difficult, is an  interesting question. Such differences in the relative difficulty of prediction tasks can be better explained by an example. Let us consider two researchers A and B attempting to predict the traffic speed for the next 5 minutes. A has access to historical traffic speed data at a resolution of 5 seconds while B has access to historical traffic speed data at a resolution of 1 second. Now, if A and B both attempt to predict the traffic speed at 5-second resolution, B's task is easier compared to A -- due to the fact that B has more information for the same task. However, if both A and B attempt to predict the traffic speed at the same resolution as their input data (A: 5 seconds; B: 1 second),  it becomes challenging to comment on whose task is easier. It should be noted that some data sources significantly reduce this problem of differences in the resolution of data. Data collected using distributed sensors (\textit{e.g.} GPS data) have a similar resolution across different countries. An empirical study on such  differences in relative difficulties of prediction tasks has been presented in \cite{ma2015large}. 
  
    \item\textit{Different lengths of prediction horizons:} Varying lengths of the prediction horizon result in different levels of complexity of the prediction task. As noted in~\cite{yu2017deep}, the performance of most models degrades when the length of the prediction horizon increases. This degradation of performance with the increase of prediction horizon is a common finding across  research~\citep{wang2016traffic, di2019traffic}. We like to highlight the obvious fact that the length of the prediction horizon and usefulness of the prediction task are inversely related. 
    
    \item\textit{Different metrics used to measure performance:} The choice of different metrics for performance poses difficulty while comparing different research. The demonstration of superior model performance using one metric does not necessarily imply superior performance when using some other metric. Congestion prediction tasks are examples of imbalanced datasets, thus the most common metrics for classification tasks (for example, accuracy) are usually misleading. In such a scenario, accuracy carries significance only when it is  reported along with other metrics carrying complementary information (for example, False Positive rates). A systematic study of the choice of metrics for classification tasks is presented in~\cite{sokolova2009systematic}.
    
    \item \textit{Differences in the heterogeneity of traffic stream}: Heterogeneous traffic stream is a common occurrence  in developing countries due to a mix of different types of vehicles and high variation in driving behaviour. Correspondingly, we observe a systematic drop in the model performance when deep learning models are used for congestion-related tasks in developing countries. We observed this trend to be valid in each of the congestion detection, prediction and alleviation sections. 
\end{enumerate}

A recent attempt to define and maintain a \textit{state-of-the-art} in various domains is \href{https://paperswithcode.com}{\textcolor{blue}{https://paperswithcode.com/sota}}. Their website presents leader boards for different tasks across various domains, including the domain of traffic prediction. This portal is slowly gaining popularity amongst the machine learning community.

\subsubsection{Lack of online learning }
Online learning (often called continuous learning) refers to the process of updating a model as new data arrives. In the absence of online learning, researchers gather all data before attempting to model the behaviour of the system. They then split the data points into three parts -- \textit{train}, \textit{validation}, and \textit{test}. The model is trained using the {train} data. In order to tune the hyper-parameters (\textit{e.g.} number of layers in the neural network), the model performance is evaluated on the validation data. Finally, after the entire training process is complete, the model performance is evaluated on the test data. The model is said to generalise well if the model performance is similar for both the validation and the test data. The approach mentioned above appears to conform to the standard practices of machine learning. However, an important caveat is that in principle, the labels for the test data should be unseen not only to the \textit{model} but also the \textit{modeller}. When the test data are available to the modeller beforehand, extreme care needs to be taken to ensure that any form of target leakage does not take place.~\cite{wujek2016best} summarise the best practices to avoid any unintentional target leakage.\\

Online learning eliminates this challenge completely. Additionally, online learning helps us model the real world scenario more closely. In the real-world deployment of a traffic prediction model, the test data are available only \textit{after} the deployment of the solution. However, we could not find many examples of research papers using models with online learning capabilities. This might be due to the security and privacy concerns with real-time traffic data being relayed to researchers. 
\subsection{Conclusions and future research directions}
This section highlights possible directions for future research. These directions are proposed in answer to the challenges which were presented in the previous subsection. Here, we highlight the importance of data standardisation, potential synergies with other domains of work and the potential synergies between the simulation-based and deep learning approaches for traffic prediction tasks. 
\subsubsection{Standardised datasets and competitions}
The success of deep learning is an example of how good quality data can transform a domain of work—the most notable example is Computer Vision (CV). CV refers to the umbrella term for tasks pertaining to the extraction of information from images. Some examples of specific CV tasks are object localisation, image classification and facial recognition. Given some constraints, the deep learning methods have achieved near-human performance levels for CV tasks. The two biggest impetuses to CV research were the release of the ImageNet \citep{imagenetwa} dataset and its use in the Pascal Visual Object recognition \citep{challengeWa}. It was noticed that the deep neural networks which had been around for a few decades, started performing very well when trained on ImageNet. \\

For the domain of traffic congestion prediction, several new possibilities arise with access to standardised datasets. First, the performance of deep learning models improves with increasing amounts of good-quality data. Second, access to common databases can allow the practitioners to establish the state-of-the-art in a disambiguous fashion. Third, it can open avenues for transfer learning where the deep neural network trained on a huge traffic dataset can be readily applied to a new city for which traffic data might not be available~\citep{tan2018survey}. Apart from these, standardisation also addresses the problem of train-validation-test split as mentioned in the previous subsection, since the {test} data are not visible to the individuals participating in the competition. 
\\
Some exclusive efforts for data standardisation in traffic prediction are \citep{NIPS2019_9213}, \citep{wang2018locality},~\citep{cheng2018deeptransport} and \citep{moosavi2019accident}. A list of the latest sources of publicly available traffic data is maintained at the~\href{https://www.polymtl.ca/wikitransport/index.php?title=Public_Transportation_Datasets}{website} of `Research in Intelligent Transportation and Road Safety at Polytechnique Montréal', though videos and images form the majority of the list. Traffic prediction tasks are also gaining popularity in the competition tracks at machine learning conferences. Some recent examples are~\href{https://www.iarai.ac.at/traffic4cast/}{traffic4cast competition at NIPS, 2020} and~\href{https://sites.google.com/view/cvpr20-scalability/competitions}{Night-owls competition at CVPR, 2020}.
\subsubsection{Similar prediction tasks in other domains}
Papers using machine learning for internet traffic classification have been popular in the Wireless Networks community for a long time. Some prediction tasks in wireless networks are similar to traffic prediction. Survey papers from the wireless network domain might hold a treasure trove of ideas for transportation researchers. Some examples of relevant surveys include~\citep{mao2018deep, zhang2019deep} and \citep{ 4738466}. A routing protocol is the counterpart of a traffic signal control strategy in transportation networks. Hence, protocol-free wireless routing is akin to adaptive signal control. Making wireless networks free from (fixed) protocols is discussed in~\cite{tang2017removing}.~\cite{yu2019deep} investigate using deep-reinforcement learning to handle the heterogeneity in wireless networks. Similarly, congestion-aware routing of wireless networks~\citep{4349670} is the counterpart of congestion-alleviation algorithms in transportation. Weather-forecasting is another domain where the prediction tasks and datasets are similar to transportation. A classic example is~\citep{xingjian2015convolutional}, which has resulted in a large number of applications in traffic~\citep{wang2020deep}. 

\subsubsection{Synergies between model-driven and deep learning based approaches}
 Deep neural networks have high predictive power and can be trained much faster and with little effort compared to calibrating a traffic simulator. Deep learning models lack {interpretability}. The inherent {interpretable} nature of traffic simulators can be leveraged by deep learning researchers when both methods are used in conjunction. Such synergies are already being explored in deep reinforcement learning-based approaches for traffic signal control. Some researchers have used microscopic traffic simulators to generate data for congestion labels, thereby solving the class imbalance problem. More such efforts in the direction of creating synergies between the two approaches are needed.
 
     
     
 
\section{Declaration of Competing Interest}
The authors declare that they have no known competing interests that could have 
influenced the research reported in this paper.

\section{Acknowledgements}
This work is an outcome of the Future Resilient Systems project at the Singapore-ETH Centre (SEC) supported by the National Research Foundation, Prime Minister’s Office, Singapore under its Campus
for Research Excellence and Technological Enterprise (CREATE) programme. The authors would like to acknowledge the valuable feedback and recommendations received from Dr. Yi Wang from Future Resilient Systems and Dr. Jimi B. Oke from University of Massachusetts, Amherst. The authors would also like to acknowledge the three anonymous reviewers for their valuable comments and suggestions.



\bibliographystyle{chicago}  
\bibliography{template.bib}  

\end{document}